\documentclass[conference]{IEEEtran}
\IEEEoverridecommandlockouts
\usepackage{cite}
\usepackage{amsmath,amssymb,amsfonts}
\usepackage{algorithmic}
\usepackage{graphicx}
\usepackage{textcomp}
\usepackage{xcolor}

\usepackage{textcomp}
\usepackage{color}
\usepackage{enumerate}
\usepackage{bm}
\usepackage{algorithm}
\usepackage{array}
\usepackage{makecell}
\usepackage{multirow}

\usepackage{enumitem,calc}
\newcommand\preitem{\mdseries\textbullet\space}
\newlist{desclist}{description}{3}
\setlist[desclist,1]{format=\preitem\bfseries,leftmargin=\widthof{\preitem},style=sameline}

\newtheorem{problem}{Problem}
\newtheorem{definition}{Definition}
\DeclareMathOperator{\softmax}{softmax}
\DeclareMathOperator*{\argmin}{arg\,min}
\newtheorem{innercustomgeneric}{\customgenericname}
\providecommand{\customgenericname}{}
\newcommand{\newcustomtheorem}[2]{%
  \newenvironment{#1}[1]
  {%
  \renewcommand\customgenericname{#2}%
  \renewcommand\theinnercustomgeneric{##1}%
  \innercustomgeneric
  }
  {\endinnercustomgeneric}
}

\newcustomtheorem{customlemma}{Lemma}

\newtheorem{proof}{Proof}

\newcommand{\dz}[1]{{\textsf{\textcolor{blue}{[dz: #1]}}}}
\newcommand{\he}[1]{{\textsf{\textcolor{red}{[From He: #1]}}}}
\newcommand{\lc}[1]{{\textsf{\textcolor{green!10!orange!90!}{[From LC: #1]}}}}
\newcommand{\hh}[1]{{\textsf{\textcolor{green}{[From HH: #1]}}}}

\newcommand{\mkclean}{
  \renewcommand{\he}[1]{}
  \renewcommand{\dz}[1]{}
  \renewcommand{\hh}[1]{}
  \renewcommand{\lc}[1]{}
}
\newcommand{\hide}[1]{}
\mkclean

\newcommand{\eg}{{\sl e.g.}}
\newcommand{\ie}{{\sl i.e.}}
\newcommand{\etc}{{\sl etc.}}

\newcommand{\name}{{\textsc{FairGen}}}

\def\BibTeX{{\rm B\kern-.05em{\sc i\kern-.025em b}\kern-.08em
    T\kern-.1667em\lower.7ex\hbox{E}\kern-.125emX}}
\begin{document}

\title{\name: Towards Fair Graph Generation}


\author{\IEEEauthorblockN{Lecheng Zheng}
\IEEEauthorblockA{\textit{Department of Computer Science} \\
\textit{University of Illinois at Urbana-Champaign}\\
Illinois, USA \\
lecheng4@illinois.edu}
\and
\IEEEauthorblockN{Dawei Zhou}
\IEEEauthorblockA{\textit{Department of Computer Science} \\
\textit{Virginia Tech}\\
Virginia, USA \\
zhoud@vt.edu}
\and
\IEEEauthorblockN{Hanghang Tong}
\IEEEauthorblockA{\textit{Department of Computer Science} \\
\textit{University of Illinois at Urbana-Champaign}\\
Illinois, USA \\
htong@illinois.edu}
\and
\IEEEauthorblockN{Jiejun Xu}
\IEEEauthorblockA{
\textit{HRL Laboratories}\\
California, USA \\
jxu@hrl.com}
\and
\IEEEauthorblockN{Yada Zhu}
\IEEEauthorblockA{
\textit{IBM}\\
Illinois, USA \\
yzhu@us.ibm.com}
\and
\IEEEauthorblockN{Jingrui He}
\IEEEauthorblockA{\textit{School of Information Science} \\
\textit{University of Illinois at Urbana-Champaign}\\
New York, USA \\
jingrui@illinois.edu}
}

\maketitle

\begin{abstract}
There have been tremendous efforts over the past decades dedicated to the generation of realistic graphs in a variety of domains, ranging from social networks to computer networks, from gene regulatory networks to online transaction networks. Despite the remarkable success, the vast majority of these works are unsupervised in nature and are typically trained to minimize the expected graph reconstruction loss, which would result in the \emph{representation disparity} issue in the generated graphs, \ie, the protected groups (often minorities) contribute less to the objective and thus suffer from systematically higher errors. In this paper, we aim to tailor graph generation to downstream mining tasks by leveraging label information and user-preferred parity constraints. In particular, we start from the investigation of representation disparity in the context of graph generative models. To mitigate the disparity, we propose a fairness-aware graph generative model named \name. Our model jointly trains a label-informed graph generation module and a fair representation learning module by progressively learning the behaviors of the protected and unprotected groups, from the `easy' concepts to the `hard' ones. In addition, we propose a generic context sampling strategy for graph generative models, which is proven to be capable of fairly capturing the contextual information of each group with a high probability. Experimental results on seven real-world data sets demonstrate that \name\ (1) obtains performance on par with state-of-the-art graph generative models across nine network properties, (2) mitigates the representation disparity issues in the generated graphs, and (3) substantially boosts the model performance by up to $17\%$ in downstream tasks via data augmentation.
\end{abstract}


\section{Introduction}
The ever-increasing size of graphs\footnote{In this paper, we use `graphs' and `networks' interchangeably.}, together with the difficulty of releasing and sharing them, has made graph generation a fundamental problem in many high-impact applications, including data augmentation~\cite{chakrabarti2006graph}, anomaly detection~\cite{akoglu2008rtm}, drug design~\cite{stokes2020deep,DBLP:conf/icml/JinBJ18}, recommendation~\cite{bojchevski2018netgan}, and many more. 
For instance, financial institutes would like to share their transaction data or user networks with their partners to improve their service. However, directly releasing the real data would result in serious privacy issues, such as the leakage of user identity information. In this case, graph generative models provide an alternative solution without privacy concerns, by generating high-quality synthetic graphs for sharing.  
The classic graph-property oriented models are usually built upon a succinct and elegant mathematical formula to preserve important structural properties, \eg, power-law degree distribution~\cite{albert2002statistical,akoglu2009rtg,leskovec2010kronecker,kim2012multiplicative}, small world phenomena~\cite{watts1998collective}, shrinking diameters of dynamic graphs ~\cite{watts1998collective, fischer2014dynamic}, local clustering~\cite{waxman1988routing, jing2023sterling, jing2022coin}, motif distributions~\cite{zhao2011synchronization}, \etc\ 
More recently, the data-driven models~\cite{bojchevski2018netgan, zheng2021deeper, DBLP:conf/cikm/ZhouZF0H22, jing2021hdmi, jing2021multiplex, you2018graphrnn,simonovsky2018graphvae,guo2020systematic,li2018learning,grover2019graphite,DBLP:conf/www/GoyalJR20} have attracted much attention, which directly extract the contextual information from the input graphs and approximate their structure distributions with minimal prior assumptions. 

Despite the tremendous success of existing graph generators, the vast majority of these generators are unsupervised and independent of downstream learning tasks. They are able to produce general-purpose graphs without considering any label information. 
However, in many real graphs, labels, such as identities of users~\cite{harrison2009identity} or community memberships~\cite{wellman1999network}, are available and could have a profound impact on the performance of downstream learning tasks. Considering an online transaction network owned by a financial institute that allows real-time money transfer among users and merchants, most of the transactions are normal while only a small number of transactions are red-flagged (\ie, fraudulent transactions) by domain experts. Such label information could play a pivotal role in financial fraud detection (\eg, money laundering detection, identity theft detection). Therefore, if the graph generators neglect such label information, it is likely to negatively impact the downstream learning tasks (\eg, fraud detection).

Moreover, as the importance of model fairness has been widely recognized in the AI community, it is highly desirable to ensure certain parity or preference constraints in the learning process of generative models~\cite{gajane2017formalizing, DBLP:conf/sdm/ZhengZH23}. In particular, it is of key importance to ensure the protected group (\eg, the African Americans) and the unprotected group (\eg, the non-African Americans) are treated equally in the generation process, especially when the generated data will be used for developing realistic AI systems (\eg, Correctional Offender Management Profiling for Alternative Sanctions (COMPAS)~\cite{DBLP:journals/corr/abs-1908-09635}). However, most, if not all, of the existing graph generative models are designed either prior to or in parallel with downstream tasks without considering model fairness in the generative process. The statistical nature of these models is designed to focus on the frequent patterns (\ie, the unprotected group), and as such, might overlook the underrepresented patterns (\ie, the protected group) in the observed data. As the protected groups contribute less to the general learning objective (\eg, minimizing the expected reconstruction loss~\cite{bojchevski2018netgan}), they tend to suffer from systematically higher errors. Following~\cite{hashimoto2018fairness}, we refer to this phenomenon as \emph{representation disparity}.
Even worse, as the protected groups are typically more scarce compared to the unprotected groups, it can be much more expensive to obtain label information from these groups than the unprotected groups in practice. As a consequence, the representation disparity issue could be further exacerbated when the models are trained with highly imbalanced label information.

Therefore, in this paper, we aim to tailor graph generation to downstream learning tasks, by incorporating both label information and parity constraints.
To this end, we have identified the following challenges.
First (\emph{C1. Task Guidance}), how to train graph generative models under the guidance of ground-truth labels, so that the generated graphs are better suited for the downstream tasks compared to the ones using general-purpose graph generators? 
Second (\emph{C2. Representation Disparity}), how to enforce the fairness constraint on the graph generative model so that the protected group and the unprotected group are treated equally in the generated graphs?
Third (\emph{C3. Label Scarcity}), given limited label information (especially for the protected groups), how to accurately capture the class memberships of the protected groups in the input data and preserve them in the generated graphs? 

To this end, we propose a deep generative model named \name, which jointly trains a label-informed graph generation module and a fair representation learning module in a mutually beneficial way. Moreover, To address the aforementioned challenges, \name\ integrates the self-paced learning paradigm in the graph generation process. It starts with few-shot labeled examples and then progressively learns the behaviors of the protected and unprotected groups, from the `easy' concepts to the `hard' ones. Moreover, to control the risk of learning protected groups, we propose a novel context sampling strategy for graph generative models, which is proven to be capable of capturing the context of each group $\mathcal{S}$ with probability at least $1 - T\delta \phi(\mathcal{S})$, where $T$ is the maximum random walk length, $\phi(\mathcal{S})$ is the conductance of subgraph $\mathcal{S}$, and $\delta$ is a positive constant. 

The main contributions of this paper are summarized below.
\vspace{-2mm}
\begin{desclist}
    \item [] \textbf{Problem.} We formalize the \emph{fair graph generation} problem and identify unique challenges motivated by real applications. 
    \item [] \textbf{Algorithms. }We propose a self-paced graph generative model named \name, which incorporates the label information and fairness constraint to produce task-specific graphs.
    \item [] \textbf{Evaluation. }We perform extensive experiments on seven real networks, which demonstrate that \name\ (1) achieves comparable performance as state-of-the-art graph generative models in terms of nine widely-used metrics; (2) largely alleviates the representation disparity in the generated graphs; (3) significantly boosts the performance of rare category detection via data augmentation. 
\end{desclist}
The rest of our paper is organized as follows. 
In Section II, we introduce our proposed method \name ~followed by experimental results in Section III. We review the related literature in Section IV before we conclude the paper in Section V.

\section{Proposed Method}
In this section, we present our \name\ framework. We first introduce the notation and the formal problem definition. Then, we provide an overview of \name\ together with its learning objective. At last, we present a graph assembling strategy for fair graph generation. 

\subsection{Notation and Problem Definition}
We formalize the graph generation problem in the context of an undirected graph $\mathcal{G} = (\mathcal{V},\mathcal{E})$, where $\mathcal{V}$ consists of $n$ vertices and $\mathcal{E}$ consists of $m$ edges. 
We let $\bm{A} \in \mathbb{R}^{n\times n}$ denote the adjacency matrix, $\bm{D} \in \mathbb{R}^{n\times n}$ denote the diagonal matrix of vertex degrees, and $\bm{I} \in \mathbb{R}^{n\times n}$ denote the identity matrix. The transition probability matrix $\bm{M}$ of $\mathcal{G}$ can be obtained by $\bm{M} = (\bm{A}\bm{D}^{-1}+\bm{I})/2$. We define an indicator vector $\bm{\chi_\mathcal{S}}\in \mathbb{R}^{n}$ which is supported on a set of nodes $\mathcal{S}\subseteq \mathcal{V}$\hide{\hh{shall we use $\subseteq$ instead?}}, \ie, $\bm{\chi_\mathcal{S}} (v) =  1$ iff $v \in \mathcal{S}$; $\bm{\chi_\mathcal{S}} (v) =  0$ otherwise.
In our problem setting, we are given a handful of labeled examples from $C$ classes and the membership of a protected group.
Without loss of generality, we let $\mathcal{L} = \{ x_1, x_2, \ldots, x_{L}\}$ denote the set of $L$ labeled vertices, which includes at least one from each class $ y = 1, \ldots, C$, 
$\mathcal{U} = \{x_{L + 1}, x_{L + 2}, \ldots, x_{L + U}\}$ denote the set of $U$ unlabeled vertices, $\mathcal{S}^+\subseteq \mathcal{V}$ denote the set of protected group vertices, and $\mathcal{S}^-\subseteq \mathcal{V}$ denote the set of unprotected group vertices. Note that $\mathcal{S}^- = \{x | x\in \mathcal{V} \ and \  x\not\in \mathcal{S}^+\}$. 
Following~\cite{bojchevski2018netgan,you2018graphrnn}, in the graph generation process, we extract $k$ random walk sequences $\mathbb{W} = \{\bm{{w}_1}, \ldots, \bm{{w}_k}\}$ from the input graph $\mathcal{G}$, where each random walk sequence consists of $T$ incident nodes traversed one after another, \ie, $\bm{{w}_i} = \{x_{i, 1}, \ldots, x_{i,T}\}$, where $x_{i, j}\in \mathcal{V}$, $j = 1, \ldots, T$. 
The learning objectives are defined to minimize the reconstruction error of generating synthetic random walks: $\bm{\widetilde{w}} \sim g_\theta(\mathbb{W})$, where $\bm{\widetilde{w}} = \{ \widetilde{x}_1, \ldots, \widetilde{x}_T\}$ is the synthetic random walk consisting of $T$ vertices $\widetilde{x}_i \in \mathcal{V}$, $i= 1,\ldots, T$, and $g_\theta$ denotes the recurrent neural network~\cite{mikolov2010recurrent,hochreiter1997long} parameterized by $\bm{\theta}$. 

\begin{figure}[t]
\begin{center}
\begin{tabular}{cc}
\includegraphics[width=0.33\linewidth]
{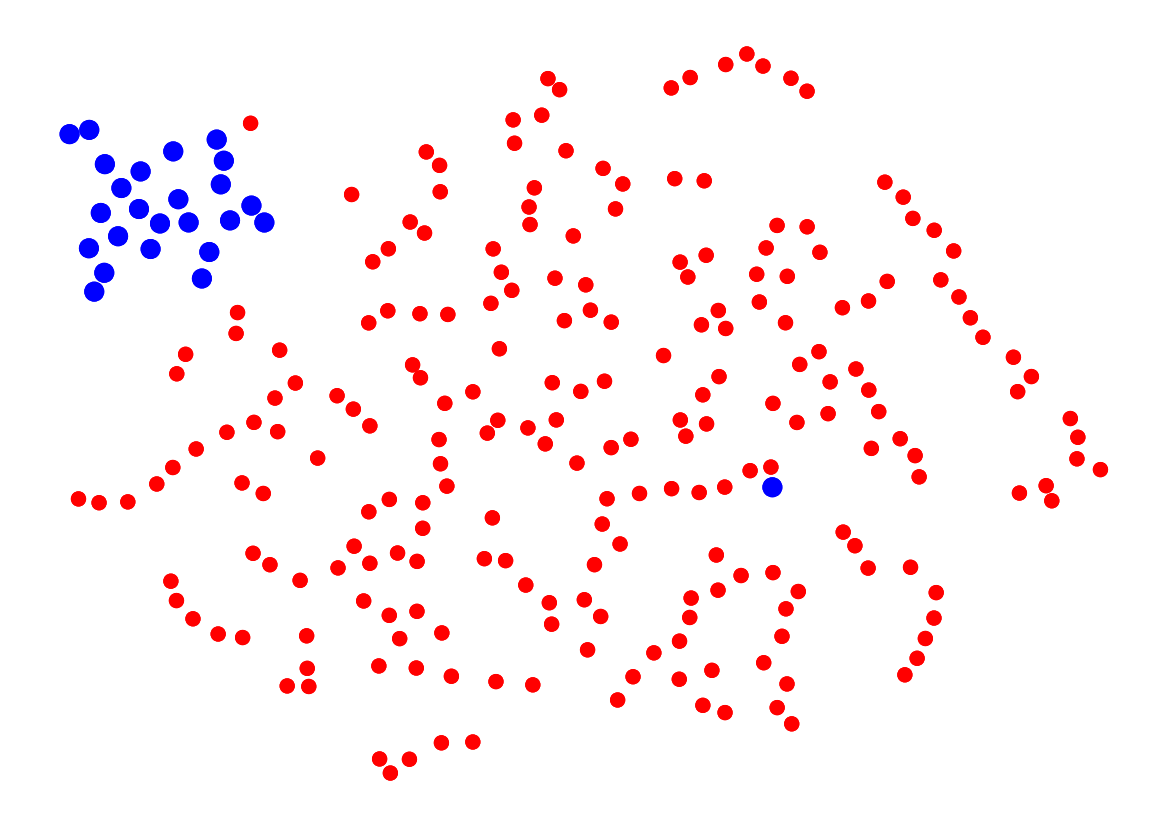}
& \includegraphics[width=0.33\linewidth]{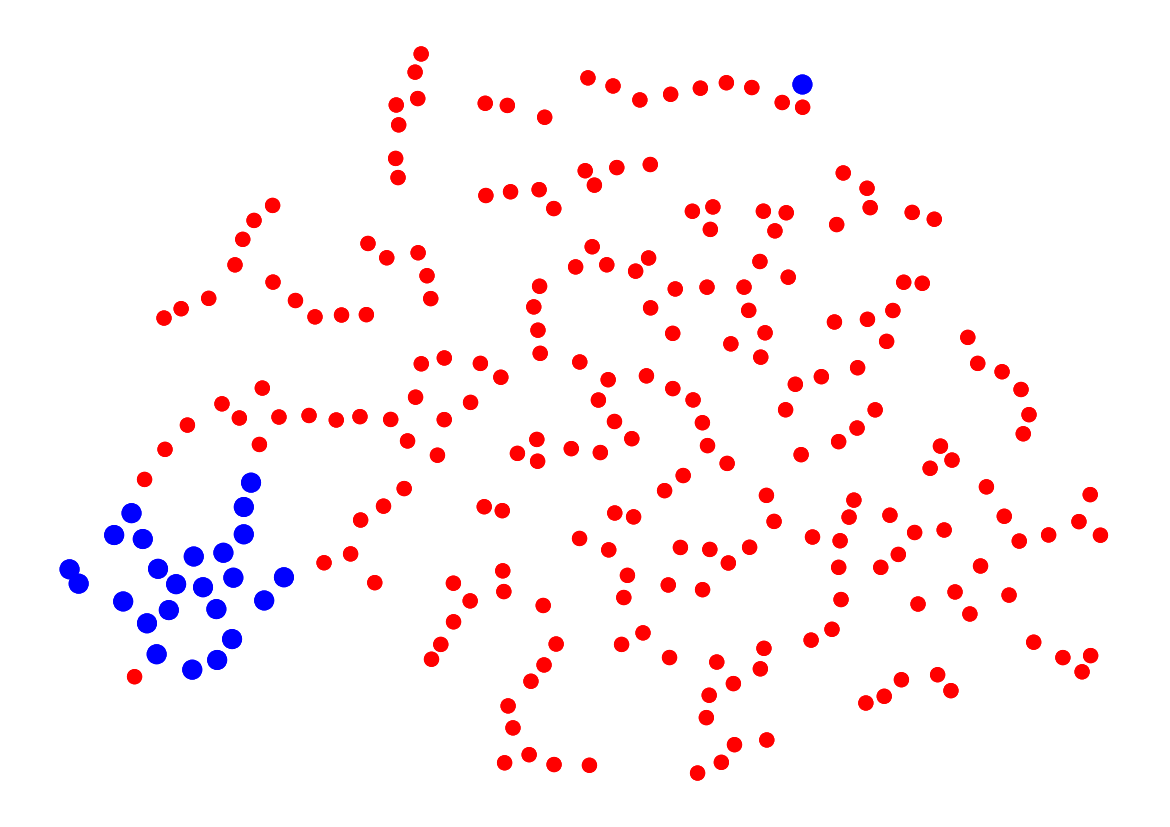}\\
(a) Original graph   & (b) \makecell[c]{NetGAN (500 iterations)} \\
\includegraphics[width=0.33\linewidth]{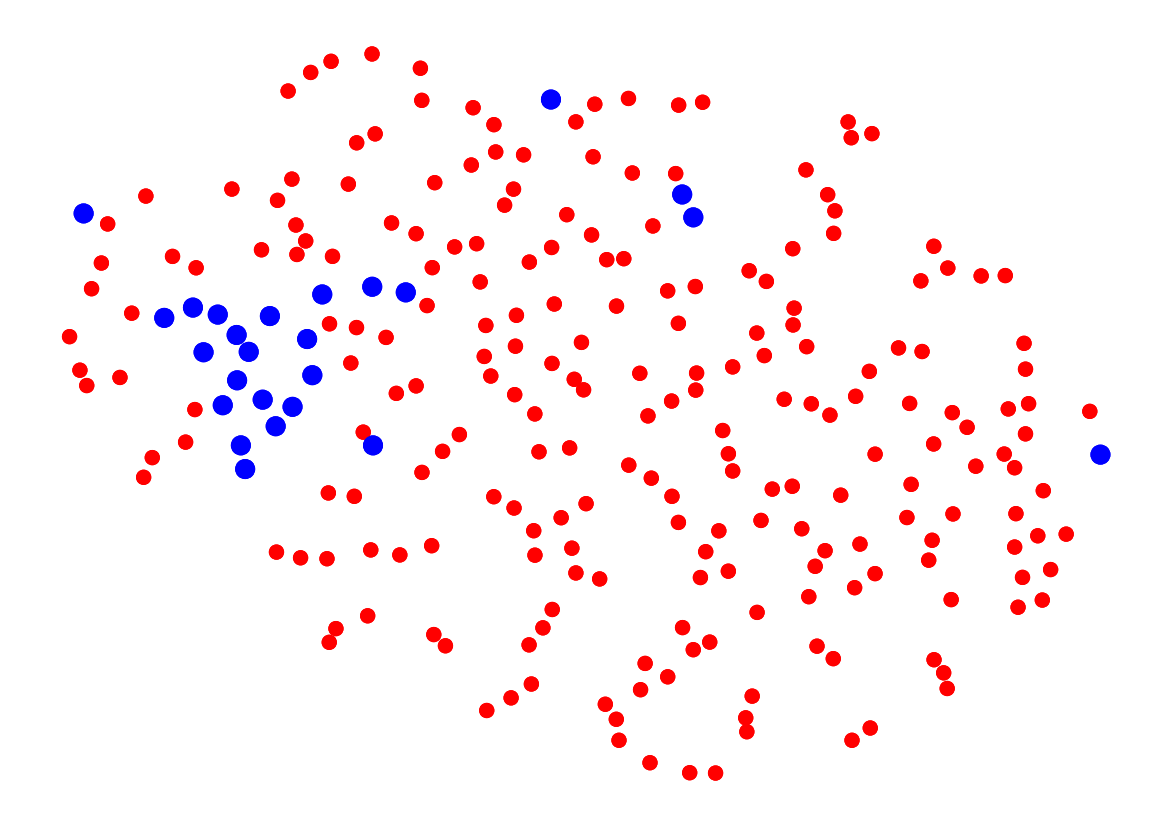}
&\includegraphics[width=0.33\linewidth]{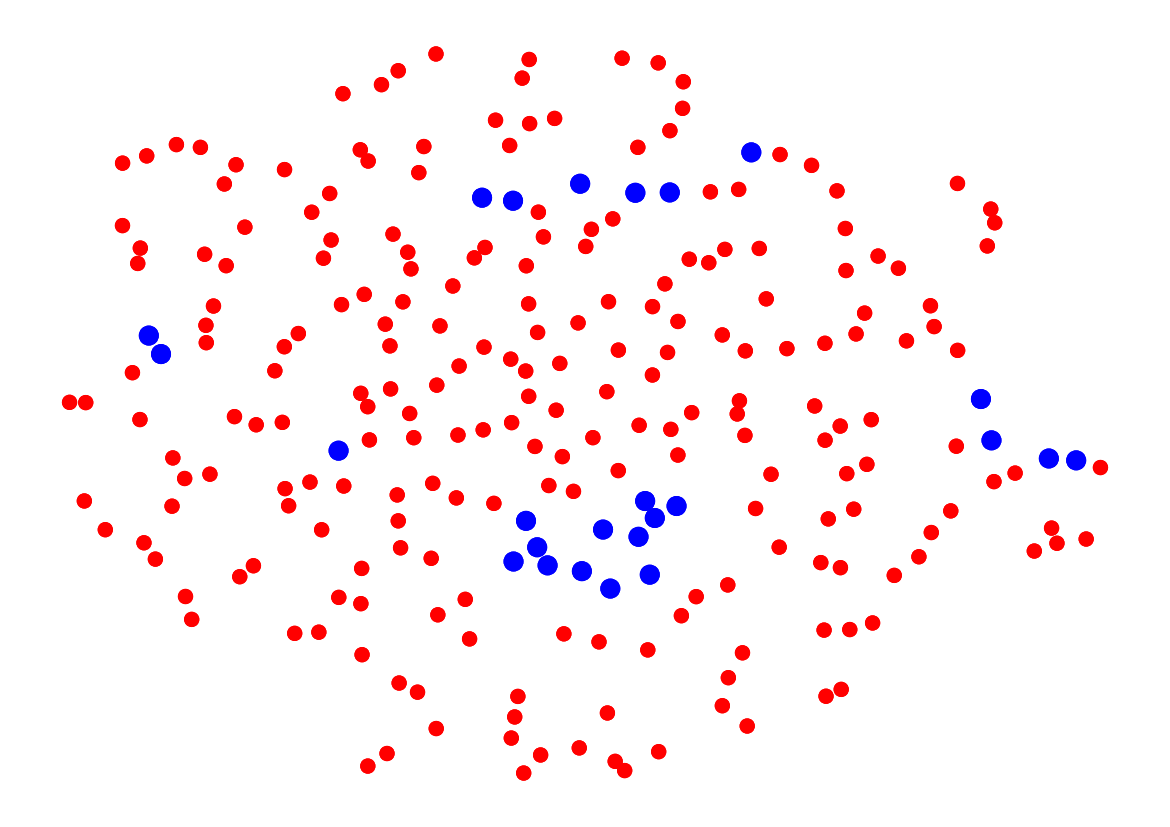} \\
(c) \makecell[c]{NetGAN (1000 iterations)}   & (d) \makecell[c]{NetGAN (2000 iterations)} \\
\end{tabular}
\end{center}
\caption{An illustrative example of representation disparity in deep graph generative models. The protected group is colored in blue while the unprotected group is colored in red. (For the purpose of better visualization, we omit the edge connection.)
}
\label{fig: disparity}
\end{figure}

\begin{figure*}[t]
\vspace{-3mm}
\includegraphics[width=\textwidth]{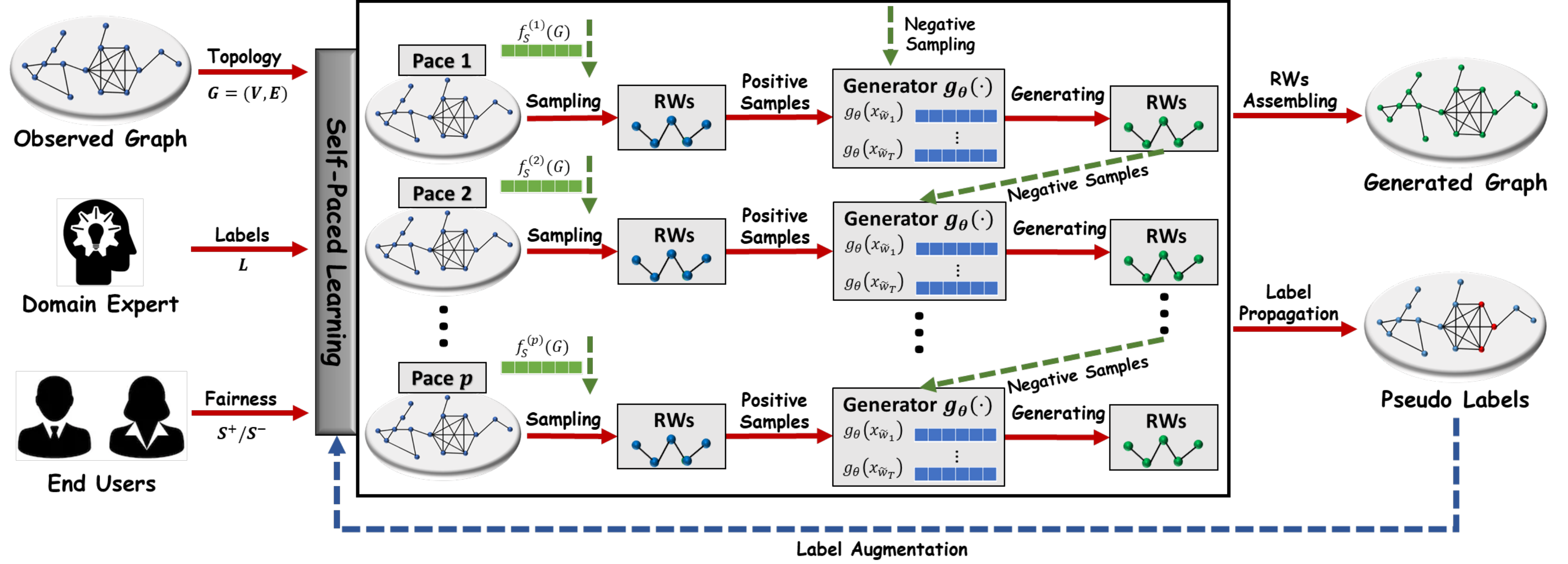}
\centering
\vspace{-1mm}
\caption{Overview of the proposed \name\ framework.
}
\label{Fig: Framework}
\end{figure*}

\textbf{Representation Disparity. }
Consider a standard graph generative model that is trained to minimize the reconstruction error of the input graph $\mathcal{G}$. 
Given the membership of the protected group $\mathcal{S}^+$, we define the general graph reconstruction loss $R (\bm{\theta})$ and the group-wise graph reconstruction loss $R_{\mathcal{S}^+}(\bm{\theta})$ as follows.
\begin{align}\label{Eq: reconstruction loss}\small
    R (\bm{\theta}) = - \mathbb{E}_{\bm{{w}} \subseteq {\mathcal{G}}}  [\sum_{t=1}^T \log g_\theta({w}_t | \bm{{w}_{<t}})]
\end{align}
\vspace{-0.3cm}
\begin{align}\label{Eq: protected reconstruction loss}\small
    R_{\mathcal{S}^+}(\bm{\theta}) = - \mathbb{E}_{\bm{{w}} \subseteq {\mathcal{G}}_{\mathcal{S}^+}}  [\sum_{t=1}^T \log g_\theta({w}_t | \bm{{w}_{<t}})]
\end{align}
where ${\mathcal{G}}_{\mathcal{S}^+}$ refers to a subgraph in  $\mathcal{G}$ that is composed of a group of vertices $\mathcal{S}^+ \subseteq {\mathcal{V}}$; ${w}_t$ and $\bm{w}_{< t}$ represent the $t^\text{th}$ node and the first $(t-1)^\text{th}$ nodes in a sampled random walk $\bm{w}$.
The objective of existing graph generative models typically aims to minimize Eq.~\ref{Eq: reconstruction loss} while ignoring the existence of the protected group $\mathcal{S}^+$ that is under-represented. However, it is vital to ensure the protected group (\eg, the African Americans) and the unprotected group (\eg, the non-African Americans) are treated equally in the generation process. 
Notice that the statistical nature of the existing graph generative models is designed to focus on the frequent patterns (\ie, the unprotected group), and as such, might overlook the underrepresented patterns (\ie, the protected group) in the observed data. Figure~\ref{fig: disparity} shows an illustrative example of this phenomenon, where we present an original graph and three synthetic graphs generated by NetGAN~\cite{bojchevski2018netgan}. We observe that the generated graphs in (b) initially maintain fairness (\ie, the protected group is well preserved in the embedding space); but as NetGAN is trained for more and more iterations, the nodes from the protected group and unprotected group get mixed together, because the protected group $\mathcal{S}^+$ contributes less to Eq.~\ref{Eq: reconstruction loss}, thus receiving less attention from the generative model (\eg, NetGAN). As a result, the status quo of generative models may obtain a low $R (\bm{\theta})$ but relatively high $R_{\mathcal{S}^+}(\bm{\theta})$. Following~\cite{hashimoto2018fairness,DBLP:conf/icml/BoseH19}, we refer to this phenomenon as the \emph{representation disparity} in graph generative models.
Here we formally define our problem below. 
{\setlength{\parindent}{0pt}
\begin{problem} \label{prob}
  \textbf{Fairness-Aware Graph Generation}\\
	\textbf{Input:} (i) an observed undirected graph $\mathcal{G}$, (ii) few-shot labeled examples $\mathcal{L} = \{x_1, \ldots, x_{L}\}$, (iii) the memberships of the protected group $\mathcal{S}^+$ and the unprotected group $\mathcal{S}^-$.\\
	\textbf{Output:} the generated graph $\widetilde{\mathcal{G}}$ that fairly preserves the contextual information (\ie, structure properties, attributes, and label information) of the protected group $\mathcal{S}^+$ and the unprotected group $\mathcal{S}^-$.
\end{problem}
}

\subsection{A Generic Joint Learning Framework}
Given a graph $\mathcal{G}$ associated with a handful of labeled nodes $\mathcal{L}$ and the membership of protected group $\mathcal{S}^+$, the goal of our framework is to learn a graph generator $g_{\bm{\theta}}$ that agrees with the known label information, and in the meanwhile fairly preserves the network context (\ie, structures and label information) of the protected group and the unprotected group in the generated graphs.
With these design objectives in mind, we formulate \name\ as an optimization problem as follows. 
\begin{align}
\argmin_{\bm{\theta}, \bm{\omega}, \bm{v}^{(1)}, \ldots, \bm{v}^{(C)}} \mathcal{J} = & \mathcal{J}_{G} + \mathcal{J}_{P} + \mathcal{J}_{F} + \mathcal{J}_{L}    + \mathcal{J}_{S} \nonumber\\
=& \underbrace{ \ \ - \ \ \mathbb{E}_{\bm{{w}}\sim f_S(\mathcal{G})} \left[\sum_{t=1}^T \log g_{\bm{\theta}} ({{w}_t} | {{\bm{w}}_{< t}}) \right]}_{\mathcal{J}_{G}\text{: label-informed generative model}}\nonumber \\
 &   \underbrace{+ \ \ \alpha \sum_{i = 1}^{L} \xi_{x_i} d_{\bm{\omega}} (x_i, y_i) }_{\mathcal{J}_{P}\text{: prediction model }} + \underbrace{\gamma \sum_{c=1}^C \|m_c^+ - m_c^-\|}_{\mathcal{J}_{F}\text{: fairness regularizer}}\nonumber
\end{align}

\begin{align}\label{Eq: objective1}\small
& \underbrace{ - \ \ \beta  \sum_{i = 1}^{L + U} \sum_{c=1}^C v_i^{(c)}  \log Pr(\hat{y}_i = c | {x}_i) }_{\mathcal{J}_{L}\text{: label propagation model}} 
 \underbrace{ - \ \ \ \lambda \sum_{i=1}^{L+U} \sum_{c=1}^C v^{(c)}_i}_{\mathcal{J}_{S}\text{: self-paced learning}}
\end{align}
where $\alpha$, $\beta$, $\gamma$, and $\lambda$ are positive constants to balance the impact of each term. The overall objective function consists of five terms.
The first term $\mathcal{J}_{G}$ corresponds to the label-informed generative model that minimizes the expected reconstruction error of the sampled random walk sequence $\bm{w}$ using the label-informed sampling strategy $f_S$.
The second term $\mathcal{J}_{P}$ minimizes the weighted prediction loss for the training data $\mathcal{L}$, where the function $\xi_{x_i}$ defines the cost-sensitive ratios regarding the protected group and the unprotected group. The third term $\mathcal{J}_{F}$ is the fairness regularizer, where $m_c^+ $ and $m_c^- $ denote the statistical parity measure~\cite{zemel2013learning} regarding the protected group $\mathcal{S}^+$ and the unprotected group $\mathcal{S}^-$, respectively. The fourth term $\mathcal{J}_{L}$ corresponds to the label propagation model that maximizes the likelihood of observing $x_i$ in its predicted class $\hat{y}_i = c$. The last term is the self-paced regularizer, which globally maintains the learning pace of graph generation and label propagation. An overview of \name\ is presented in Figure~\ref{Fig: Framework}. It consists of three major components, including (M1) label-informed graph generative module (\ie, $\mathcal{J}_{G}$), (M2) fair learning module (\ie, $\mathcal{J}_{P} + \mathcal{J}_{F}$), and (M3) self-paced learning module (\ie, $\mathcal{J}_{L} + \mathcal{J}_{S}$). Next, we will elaborate on these components one by one.

\begin{figure}[t]
\includegraphics[width=0.5\textwidth]{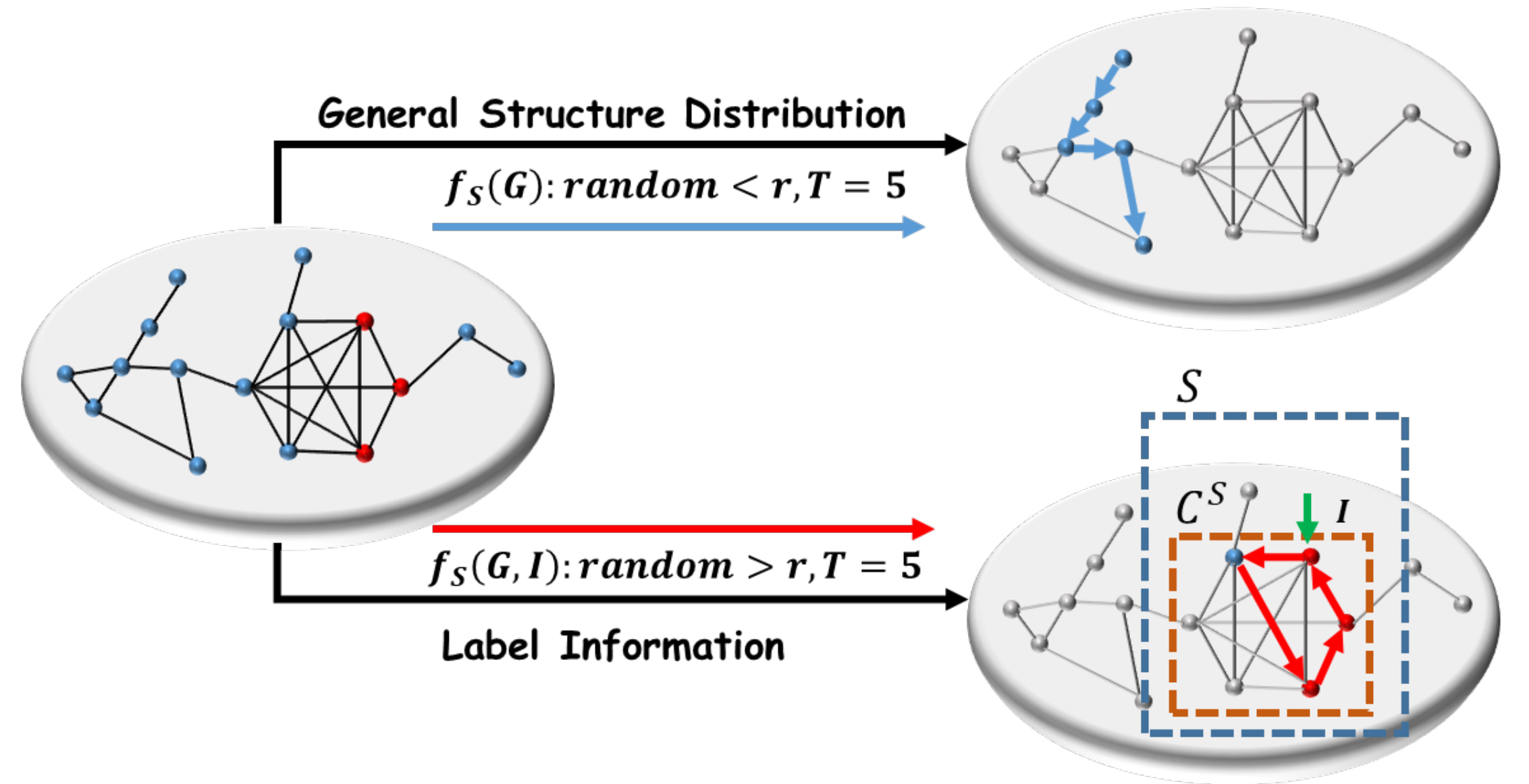}
\centering
\caption{An illustrative example of random walk extraction via $f_S(\cdot)$, where the red dots represent the labeled examples, and the blue dots represent the unlabeled examples. With probability $r$, $f_S(\cdot)$ samples random walks (colored blue) for capturing general structure distribution; with probability $1-r$, $f_S(\cdot)$ samples random walks (colored red) starting from a labeled example (pointed by a green arrow).}
\label{Fig: sampling}
\end{figure}

\textbf{M1. Label-informed graph generative module:}
The existing RNN-based graph generative models~\cite{bojchevski2018netgan,you2018graphrnn} often suffer from the long training process when modeling large-scale networks. To reduce the time complexity, we aim to train the transformer-based generator~\cite{vaswani2017attention} to model long symbolic sequences as follows.
\begin{align}\label{Eq: objective-gen}
\small
    \arg\min_{\bm{\theta}}  - \mathbb{E}_{\bm{{w}}\sim f_S(\mathcal{G})} \left[\sum_{t=1}^T \log g_{\bm{\theta}} ({{w}_t} | {{\bm{w}}_{< t}}) \right]
\end{align}
where $g_{\bm{\theta}} $ is the Transformer-based generator; $f_S(\cdot)$ is a label-informed context sampling function; ${w}_t$ and $\bm{w}_{< t}$ represent the $t^\text{th}$ node and the first $(t-1)^\text{th}$ nodes in a specific random walk $\bm{w}$. However, one major drawback of most graph generative models is the representation disparity. To alleviate this issue, we propose $\mathcal{J}_{G}$ to approximately minimize $R_{\mathcal{S}^+} (\bm{\theta})$ in Eq.~\ref{Eq: protected reconstruction loss} and $R(\bm{\theta})$ in Eq.~\ref{Eq: reconstruction loss} across both protected group (\ie, $\mathcal{S} = \mathcal{S}^+$) and unprotected group (\ie, $\mathcal{S} = \mathcal{S}^-$)
\he{Why is the unprotected group involved in minimizing $R_{\mathcal{S}^+} (\bm{\theta})$?} 
via $f_S(\cdot)$. In particular, $f_S(\cdot)$ is designed to extract two types of context information from the input data. The first type of context is based on the graph $\mathcal{G}$, which encodes the general structure distribution by minimizing $R (\bm{\theta})$ in Eq.~\ref{Eq: reconstruction loss}. The second type of context is based on the labeled examples, which encode the class-membership information. 
In Figure~\ref{Fig: sampling}, we present an example of extracting two types of random walks via $f_S(\cdot)$ on a toy graph. In this figure, we can see label-informed random walks (colored in red) traverse within the subgraph $\mathcal{S}$ (bounded by the blue box), by starting from a labeled example (indicated by the green arrow) in $C^\mathcal{S}$ (\ie, the clique bounded by the orange box).
Without loss of generality, we assume that all the labeled examples are representative, \ie, located within the diffusion cores~\cite{apers2019expansion,spielman2013local} of the corresponding classes, as defined below.

\begin{definition}\textbf{[Diffusion Core]}\label{DF: Diffusion Core}
\textit{For any subgraph $\mathcal{S}\subseteq \mathcal{G}$, the $(\delta, t)$-diffusion core of $\mathcal{S}$ is defined as 
$ C^\mathcal{S} = \{ x \in \mathcal{S} | 1 - \bm{\chi}_{\mathcal{S}} M^t \bm{\chi}_x < \delta \phi(\mathcal{S})\}$, where $\delta \in( 0, 1)$, $M = (AD^{-1}+I)/2$ is the transition probability matrix, $\bm{\chi_\mathcal{S}}$ and $\bm{\chi_x}$ are two indicator vectors supported on $\mathcal{S}$ and $\{x\}$, and $\phi(\mathcal{S})$ denotes the conductance of $\mathcal{S}$ in $\mathcal{G}$.}
\end{definition}

Note that $1 - \bm{\chi}_{\mathcal{S}} M^t \bm{\chi}_x$ computes the probability of a random walk starting from node $x\in \mathcal{S}$ and escaping $\mathcal{S}$ after $t$ steps. Roughly speaking, $C^\mathcal{S}$ is the set of nodes that are connected with each other within the subgraph $\mathcal{S}$. Next, in Lemma~\ref{lemma_sampling}, we show that if the labeled example is located in the diffusion core of $\mathcal{S}$, the extracted random walk sequences will purely preserve the context information within $\mathcal{S}$ with a high probability. 

\begin{customlemma}{2.1}\label{lemma_sampling}
\textit{If the labeled example $x_i$ is located in the diffusion core of a subgraph $\mathcal{S}$, \ie, $x_i \in C^\mathcal{S}$, then the sampled $T$-length random walks starting from $x_i$ only capture the context information within $\mathcal{S}$ with probability at least $1 - T\delta \phi(\mathcal{S})$.}
\end{customlemma}


\begin{proof}
\label{Lemma_proof}
To ensure the sampled random walks $\bm{w}$ only preserve the context information of $\mathcal{S}$, we need $\bm{w}$ to stay entirely inside of $\mathcal{S}$. Note that $\bm{M} \bm{\chi_x}$ is the distribution mass that a one-step random walk starts from $x_i$ and $diag(\bm{\chi_\mathcal{S}})\bm{M} \bm{\chi_x}$ is the truncated distribution when the $\bm{w}$ stays inside $\mathcal{S}$. Therefore, the probability of the extracted $T$-length random walks entirely staying inside of subgraph $\mathcal{S}$ is $\bm{1}' (diag(\bm{\chi_\mathcal{S}})\bm{M})^t \bm{\chi_x}$. 

For any $1 \leq t \leq T$, we can have 
\begin{align}
    &\bm{1}' (diag(\bm{\chi_\mathcal{S}})\bm{M})^{t-1} \bm{\chi_x} - \bm{1}' (diag(\bm{\chi_\mathcal{S}})\bm{M})^{t} \bm{\chi_x}\\
    &=\bm{1}' (I - diag(\bm{\chi_\mathcal{S}})\bm{M})(diag(\bm{\chi_\mathcal{S}})\bm{M})^{t-1} \bm{\chi_x}\nonumber\\
    &=\bm{1}' (\bm{M} - diag(\bm{\chi_\mathcal{S}})\bm{M})(diag(\bm{\chi_\mathcal{S}})\bm{M})^{t-1} \bm{\chi_x}\nonumber\\
    &=\bm{1}' (I - diag(\bm{\chi_\mathcal{S}}))\bm{M}(diag(\bm{\chi_\mathcal{S}})\bm{M})^{t-1} \bm{\chi_x}\nonumber\\
    &=\bm{\chi_{\bar{\mathcal{S}}}}\bm{M}(diag(\bm{\chi_\mathcal{S}})\bm{M})^{t-1} \bm{\chi_x}\nonumber\\
    &\leq \delta \phi(\mathcal{S})\bm{\chi_{\bar{\mathcal{S}}}}\bm{M}^{t} \bm{\chi_x}\nonumber
\end{align}

Based on Def.~\ref{DF: Diffusion Core}, we have 
\begin{align}\label{Eq: lemma1}
    \bm{1}' (diag(\bm{\chi_\mathcal{S}})\bm{M})^{t-1} \bm{\chi_x} - \bm{1}' (diag(\bm{\chi_\mathcal{S}})\bm{M})^{t} \bm{\chi_x} \leq \delta \phi(\mathcal{S})
\end{align}

For $t = 1, \ldots, T$, the Eq.~\ref{Eq: lemma1} can be written as follows. 
$$ 1 - \bm{1}' (diag(\bm{\chi_\mathcal{S}})\bm{M})^{1} \bm{\chi_x} \leq \delta \phi(\mathcal{S})$$

$$ \bm{1}' (diag(\bm{\chi_\mathcal{S}})\bm{M})^{1} \bm{\chi_x} - \bm{1}' (diag(\bm{\chi_\mathcal{S}})\bm{M})^{2} \bm{\chi_x} \leq \delta \phi(\mathcal{S})$$
$$ \vdots $$
$$ \bm{1}' (diag(\bm{\chi_\mathcal{S}})\bm{M})^{T-1} \bm{\chi_x} - \bm{1}' (diag(\bm{\chi_\mathcal{S}})\bm{M})^{T} \bm{\chi_x} \leq \delta \phi(\mathcal{S})$$

By adding up the above $T$ inequalities, we have 
\begin{align}
    1 - \bm{1}' (diag(\bm{\chi_\mathcal{S}})\bm{M})^{T} \bm{\chi_x} \leq T \delta \phi(\mathcal{S})\textbf{}
\end{align}

Thus, we have proved that $\bm{w}$ only preserves the context information of $\mathcal{S}$ with the probability of $\bm{1}' (diag(\bm{\chi_\mathcal{S}})\bm{M})^{T} \geq 1 - T \delta \phi(\mathcal{S}) $.
\end{proof}

In practice, we want to control $\mathcal{S}$ to be compact, such that (1) $\phi(\mathcal{S})$ is small and $1 - T\delta \phi(\mathcal{S})$ is close to 1; (2) the extracted group-wise contextual information is meaningful. 
We describe the technical details of $f_S(\cdot)$ as follows.
We first sample a random number $r\in[0,1]$. Then, with probability $r$, we uniformly sample a $T$-length random walk $\bm{w}$ via the biased second-order random walk sampling strategy~\cite{grover2016node2vec}; with probability $1-r$, we sample graph context with the guidance of label information. 

\textbf{M2. Fair learning module:} 
Through M1, we encode the general structure distribution and the label information of the input data into the graph generator $g_{\bm{\theta}}$ via $f_S(\cdot)$. Nevertheless, simply minimizing the reconstruction loss defined in Eq.~\ref{Eq: objective-gen} may overlook the protected group nodes, due to the imbalanced nature between the protected and unprotected groups. To minimize the risk of representation disparity in $g_{\bm{\theta}}$, we propose a self-paced label propagation to gradually generate `\emph{accurate and fair}' labels to be fed to $f_S(\cdot)$ for label-informed context sampling. In particular, given a handful of labeled examples together with the membership of the protected group, the learning objective of this module is to minimize the following two terms (\ie, $\mathcal{J}_{P} + \mathcal{J}_{F}$) below. 
\begin{align}\label{Eq: M2}
\small
     & \ \ \alpha \sum_{i = 1}^{L} \xi_{x_i} d_{\bm{\omega}} (x_i, y_i)   + \gamma \sum_{c=1}^C \|m_c^+ - m_c^-\|  
\end{align}
where $d_{\bm{\omega}}$ is the discriminator that learns the mapping between $x_i$ and $y_i$ via cross-entropy loss. The architecture of the discriminator is a three-layer MLP.
In Eq. \ref{Eq: M2}, the function $\xi_{x_i}$ defines the cost-sensitive ratios regarding the protected group and the unprotected group as follows.
\begin{align}\label{Eq: cost-sensitive}\small
\xi_{x_i} =
\begin{cases}
1/|\mathcal{S}^+|& \text{$x_i \in \mathcal{S}^+$}\\
1/|\mathcal{S}^-|& \text{Otherwise.}
\end{cases}
\end{align}
where $|\mathcal{S}^+|$ ($|\mathcal{S}^-|$) denotes the cardinality of $\mathcal{S}^+$ ($\mathcal{S}^-$).
Intuitively, as the protected group often corresponds to the minorities, \ie, $|\mathcal{S}^+| \ll |\mathcal{S}^-|$, we have $\xi_{x_i} \gg \xi_{x_j}$ for $x_i\in \mathcal{S}^+$ and $x_j\in \mathcal{S}^-$. By enforcing a higher loss of misclassifying protected group nodes, the predictor $d_{\bm{\omega}} $ tends to pay more attention to the underrepresented protected group $\mathcal{S}^+$. The second term guarantees the label propagation is `\emph{fair}' via statistical parity constraint~\cite{zemel2013learning}, where 
\begin{align}\small
\label{Eq: parity+}
&m_c^+ = \frac{1}{|\mathcal{S}^+|}\sum_{{x}_i \in \mathcal{S}^+} \log Pr(\hat{y}_i = c| {x}_i)
\end{align}
\begin{align}\small
&m_c^- = \frac{1}{|\mathcal{S}^-|}\sum_{{x}_j \in \mathcal{S}^-} \log Pr(\hat{y}_j = c| {x}_j)\label{Eq: parity-}
\end{align}
where $Pr(y_i | x_i)= \softmax(\bm{h}(x_i))$ and $\bm{h}(x_i)$ is the hypothesis learned by $d_{\bm{\omega}}$ mapping node $x_i$ to the label space $y_i$. We aim to ensure that the label propagation is `\emph{accurate}' by maximizing the likelihood probability $Pr(\hat{y}_i = c | {x}_i)$. Intuitively, we would like to ensure the expected probability of a protected group node $x_i \in \mathcal{S}^+$ from a particular class $\hat{y}_i = c$ is close to the expected probability of an unprotected group node $x_j \in \mathcal{S}^-$ belonging to the same class $\hat{y}_j=c$. For example, in a professional network, we want to ensure the female programmers (protected group $\mathcal{S}^+$) have the same chance to be promoted to the position of the principal scientist as the male programmers (unprotected group $\mathcal{S}^-$) in an IT company. As shown in Figure~\ref{Fig: Framework}, in each iteration, \name\ feeds the generated pseudo labels and the ground truth labels to $f_S(\cdot)$ for training $g_{\bm{\theta}} $ via negative sampling~\cite{mikolov2013distributed,mikolov2013efficient}.

\textbf{M3. Self-paced learning module:}
Though we could generate a graph by preserving the structural properties of both the protected group and unprotected group with M1 and M2, the generative model may still suffer from the issue of label scarcity. As mentioned earlier,  the protected groups are typically more scarce compared to the unprotected groups, and thus, it can be much more expensive to obtain label information from these groups than the unprotected groups in practice. To address this issue, we propose to regularize the learning process via a self-paced label module, which aims to minimize the following two terms (\ie, $\mathcal{J}_{L} + \mathcal{J}_{S}$).
\begin{align}\label{self-paced_learning}
-\beta  \sum_{i = 1}^{L + U} \sum_{c=1}^C v_i^{(c)}  \log Pr(\hat{y}_i = c | {x}_i) - \lambda \sum_{i=1}^{L+U} \sum_{c=1}^C v^{(c)}_i
\end{align}
where $\bm{v}^{(c)}\in \{0,1\}^{n\times 1}$ denotes the self-paced vectors regarding the class $c= 1, \ldots, C$. The general philosophy of self-paced learning~\cite{kumar2010self} is to learn from the `easy' concepts to the `hard' ones following the cognitive mechanism of human beings. The purpose of Eq. \ref{self-paced_learning} is to globally maintain the learning pace of the graph context extraction in {M1} and the label propagation in {M2} such that the two modules are trained in a mutually beneficial way. To be specific, at each self-paced cycle $l =1, \ldots, p$ shown in Figure~\ref{Fig: Framework}, {M3} first computes the self-paced vectors $\bm{v}^{(c)}$, $c= 1, \ldots, C$, to assign pseudo labels to a set of unlabeled vertices using the self-paced threshold $\lambda$ and the learned predictive model $d_{\bm{\omega}}$ in the last cycle $l-1$; then {M1} samples new random walks based on the updated self-paced vectors $\bm{v}^{(c)}$ and updates the generative model in Eq.~\ref{Eq: objective-gen} via negative sampling~\cite{grover2016node2vec}. In particular, at each cycle $l$, we treat the newly sampled random walks via $f_S(\cdot)$ as positive samples and the generated random walks from last cycle $l-1$ as negative samples. In this way, we gradually increase the learning difficulty of $g_{\bm{\theta}}$ and force it to distinguish the characteristics of the real random walks from the fake ones, in order to better model the distributions of the protected and the unprotected groups. In the meanwhile, {M2} updates the predictive model by learning from the augmented training data (\ie, labeled data and pseudo labeled data) that is preserved in the updated self-paced vectors $\bm{v}^{(c)}$.

Mathematically, the self-paced vectors $\bm{v}^{(c)}$ serve as a key component for training M1 and M2. In particular, we gradually increase the value of $\lambda$ for increasing the learning difficulty, which will be used to update the self-paced vectors in the next learning cycle. By taking the partial derivative of $\mathcal{J}$ in Eq.~\ref{Eq: objective1}, the gradient of $\bm{v}^{(c)}$ can be written as follows.
\begin{equation}
\small
\begin{split}
\frac{\partial \mathcal{J}}{\partial v^{(c)}_i} = - \log Pr(\hat{y_i} = c| \bm{x_i}) - \lambda 
\end{split}
\end{equation}
Thus, the closed-form solution of updating $v^{(c)}_i$ is  
\begin{equation}\label{Eq: Update V}
\small
\begin{split}
v^{(c)}_i =
\begin{cases}
1& \text{$-  \log Pr(\hat{y_i} = c| {x_i}) <  \lambda $}\\
0& \text{Otherwise}
\end{cases}
\end{split}
\end{equation}
Intuitively, $\lambda$ serves as a learning threshold for selecting the nodes to be labeled. 
In particular, when $v_i^{(c)} = 1$, it indicates \name\ classifies $x_i$ to class $c$ with a high confidence $\log Pr(\hat{y_i} = c| {x_i}) >  - \lambda$; when $v_i^{(c)} = 0$, it indicates the prediction loss $- \log Pr(\hat{y_i} = c| {x_i})$ is higher than the threshold $\lambda$. By monitoring the increased rate of $\lambda$ over self-paced cycles $l =1, \ldots, p$, the end users can easily control the learning pace and learning difficulty. 
In fact, \name\ propagates the pseudo labels to the unlabeled vertices from the easy concept (\ie, the ones with a small loss $-  \log Pr(\hat{y_i} = 1| \bm{x_i})$) to the hard ones (\ie, the ones with a large loss $-  \log Pr(\hat{y_i} = 1| \bm{x_i})$) by gradually increasing the value of $\lambda$.

\begin{algorithm}[t]\small
\caption{The \name\ Learning Framework.}
\label{Alg2}
\begin{algorithmic}[1]
\REQUIRE ~~\\
    (i) an undirected graph $\mathcal{G}$, (ii) few-shot labeled examples $\mathcal{L} = \{x_1, \ldots, x_{|L|}\}$, (iii) the membership of the protected group vertices $\mathcal{S}^+$. (iv) parameters $T$, $K$, $T_1$, $N_1$, $p$, $r$, $\alpha$, $\beta$, $\gamma$, $\lambda$.
\ENSURE ~~\\
    Generative model $g_{\bm{\theta}}$, predictive model $d_{\bm{\omega}} $,  self-paced vectors $\bm{v}^{(1)}, \ldots, \bm{v}^{(C)}$\\
    \STATE Initialize the predictive model $d_{\bm{\omega}} $ and the self-paced vectors $\bm{v}^{(1)}, \ldots, \bm{v}^{(C)}$ based on the labeled vertices $\mathcal{L}$.
    \STATE Sample $K$ positive random walks  via $f_S$ and store them in $\mathcal{N}^+$; sample $K$ negative random walks based on~\cite{grover2016node2vec} and store them in $\mathcal{N}^-$.
    \FOR{$l = 1, \ldots, p$}
        \STATE Update the hidden parameters $\bm{\theta}$ of the generator $g_{\bm{\theta}}$ by training from $\mathcal{N}^+$ and $\mathcal{N}^-$. 
        \STATE Sample $K$ positive random walks by $f_S$ with the updated self-paced vectors $\bm{v}^{(1)}, \ldots, \bm{v}^{(C)}$ and add them to $\mathcal{N}^+$.
        \STATE Sample $K$ negative random walks using the current generative model $g_{\bm{\theta}}$ and add them to $\mathcal{N}^-$. 
        \STATE Augment the value of $\lambda$.
        \STATE Update $\bm{v}^{(1)}, \ldots, \bm{v}^{(C)}$ based on Eq.~\ref{Eq: Update V} and augment $\mathcal{L}$ with the pseudo labeled vertices.
        \FOR{$t = 1: T_1$}
            \STATE Sample $N_1$ labeled vertices from $\mathcal{L}$ and update hidden layers' parameters $\bm{\omega}$ by taking a gradient step with respect to $\mathcal{J}_{P} + \mathcal{J}_{L} + \mathcal{J}_{F}$.
        \ENDFOR
    \ENDFOR
\end{algorithmic}
\end{algorithm}

\subsection{Optimization Algorithm}
\label{Optimization}
To optimize the overall objective function described in Eq.~\ref{Eq: objective1}, we present the optimization algorithm in Algorithm~\ref{Alg2} for learning \name\ framework. The inputs include an undirected graph $\mathcal{G}$ together with the labeled vertices $\mathcal{L}$, the memberships of the protected group $\mathcal{S}^+$, the length of random walks $T$, the number of sampled random walks $K$, batch iterations $T_1$, batch size $N_1$, the number of self-paced cycles $p$ and parameters $r$, $\alpha$, $\beta$, $\gamma$, $\lambda$. In Step 1, we first initialize the predictive model $d_{\bm{\omega}} (\cdot)$ and the self-paced vectors $\bm{v}^{(1)}, \ldots, \bm{v}^{(C)}$ based on the labeled vertices $\mathcal{L}$. Specifically, we let $\bm{v}^{(c)}_i =1$ for all the vertices $x_i$ labeled as class $c$; otherwise, $\bm{v}^{(c)}_i =0$. Step 2 samples $K$ positive random walks and $K$ negative random walks and stores them in $\mathcal{N}^+$ and $\mathcal{N}^-$ respectively. Step 3 to Step 12 is the main body of Algorithm 2. In particular, at each self-paced cycle $l= 1,\ldots, p$, Step 4 updates the generative model $g_\theta (\cdot)$ by learning from $\mathcal{N}^+$ and $\mathcal{N}^-$. Step 5 and Step 6 sample new positive random walks and negative random walks for training $g_\theta (\cdot)$ in the next cycle $l+1$. By adding the generated random walks to $\mathcal{N}^-$, we are increasing the difficulty of training $g_\theta (\cdot)$. In this way, we enforce $g_\theta (\cdot)$ to distinguish the characteristics of the real random walks from the fake ones and then generate better random walks that are plausible in the real graph. Step 7 and step 8 update the self-paced vectors and $\lambda$, which will be used to augment the training set $\mathcal{L}$ with the pseudo-labeled vertices. At last, from Step 9 to Step 11, we employ stochastic gradient descent (SGD)~\cite{bottou2010large} to minimize the objective function of \emph{M2}.

\subsection{Fair Network (\name) Assembling}\label{assembling}
After obtaining $g_{\bm{\theta}} $ and $d_{\bm{\omega}} $, we construct a score matrix $\bm{B}\in \mathbb{R}^{n \times n}$ to infer the adjacency matrix $\widetilde{\bm{A}}$ of the output graph $\widetilde{\mathcal{G}}$.
In particular, we let the learned generative model $g_{\bm{\theta}} $ continuously generate synthetic random walks $\widetilde{\bm{w}}$, and then collect the counts of each observed edge $(i,j)$ to be stored in $\bm{B}(i,j)$. However, simply thresholding $\bm{B}$ to produce $\widetilde{\bm{A}}$ may lead to the low-degree nodes or protected group nodes being left out. Here, we propose the following assembling criteria: (1) the protected group $\mathcal{S}^+$ in the generated graph $\widetilde{\mathcal{G}}$ should have a similar volume (total number of edges) as the original graph $\mathcal{G}$; (2) each node should have at least one connected edge in the generated graph $\widetilde{\mathcal{G}}$. Typically, we generate a much larger number of random walks than the sampled ones, which is beneficial to ensure the overall quality and to reduce the randomness of the generated graphs. Finally, we threshold $\bm{B}$ to produce $\widetilde{\bm{A}}$, which has the same number of edges as in $\bm{A}$.

\section{Experiments}
We demonstrate the performance of \name\ on seven real-world graphs with respect to graph generation, data augmentation, parameter sensitivity, and scalability. 

\begin{table}[t]
\centering  
\begin{tabular}{|c|c|c|c|c|c|}
\hline Category  &  Network &  Nodes  &  Edges & Class & \makecell{Protected \\ Group}\\
\hline
\hline Communication   &  Email  &  1,005  &  25,571  & N/A&N/A\\
\hline \multirow{3}{*}{Social Network}  &  FB  &  4,039  &  88,234  & N/A& N/A\\
                & BLOG & 5,196 & 360,166 & 6 & 300\\
                & FLICKR & 7,575 & 501,983 & 9 & 450\\
\hline File-Sharing   &  GNU  &  6,301  &  20,777 & N/A& N/A \\
\hline \multirow{2}{*}{Collaboration}   &  CA  &  5,242  &  14,496 & N/A& N/A\\
                              & ACM & 16,484 & 197,560 & 9 & 597\\
\hline
\end{tabular}
\caption{Statistics of the datasets.}
\label{TB:Net}
\end{table}

\begin{figure*}[t]
\begin{center}
\begin{tabular}{ccc}
\includegraphics[width=0.316\linewidth]{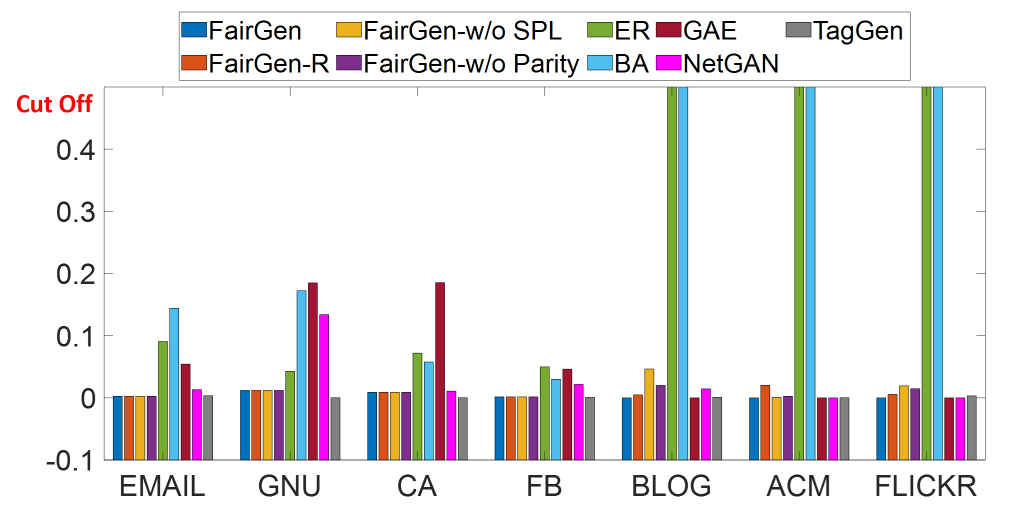} 
&\includegraphics[width=0.316\linewidth]{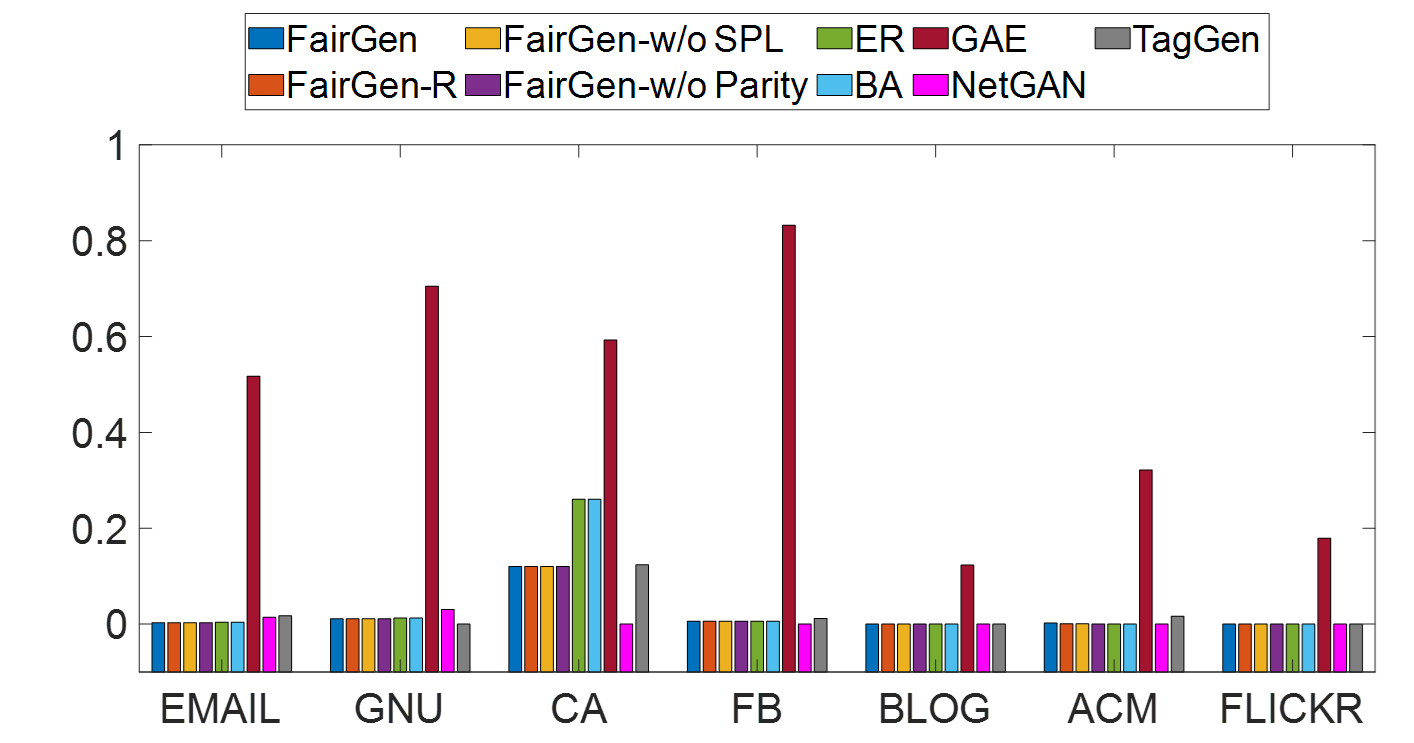}
&\includegraphics[width=0.316\linewidth]{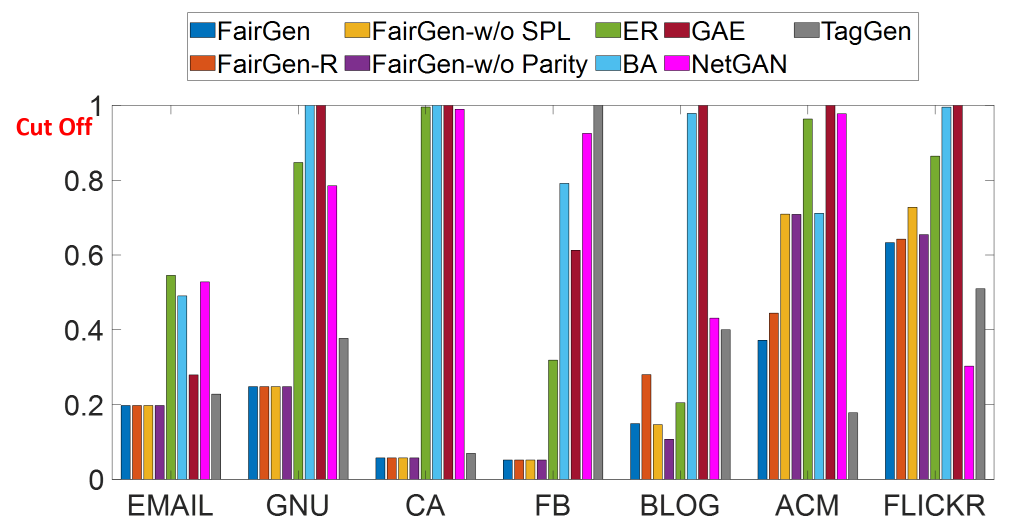} \\
(a) Average Degree  & (b)  LLC & (c) Triangle Count \\
\includegraphics[width=0.316\linewidth]{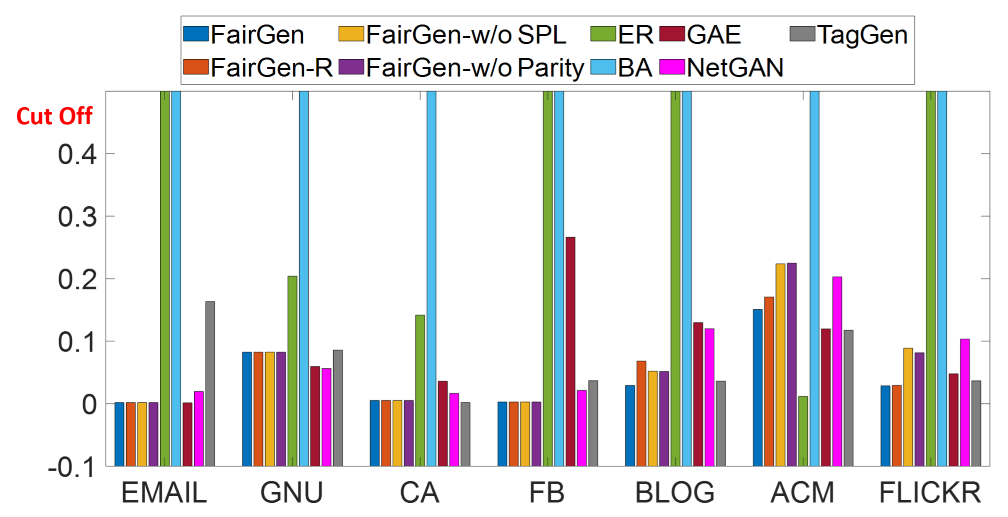}
&\includegraphics[width=0.316\linewidth]{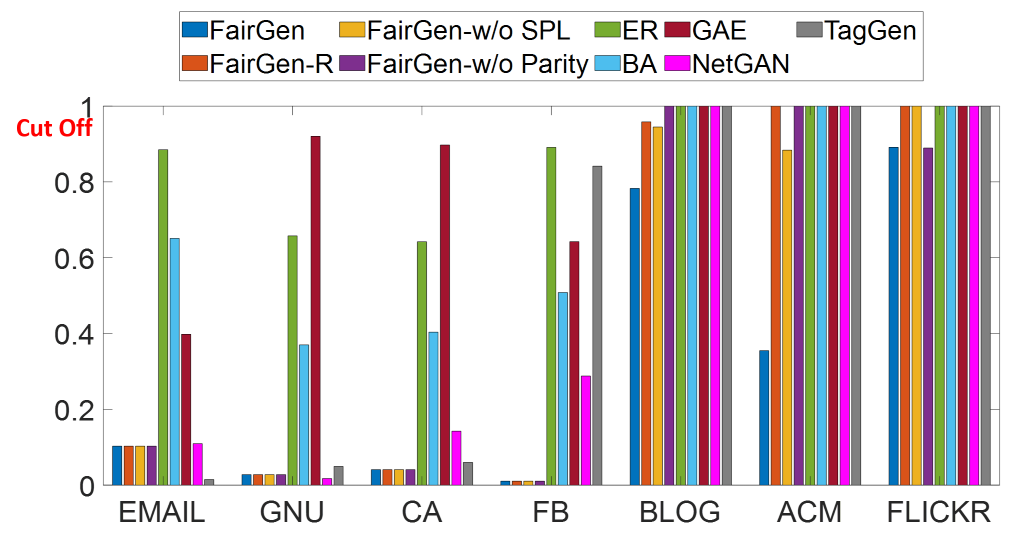}
&\includegraphics[width=0.316\linewidth]{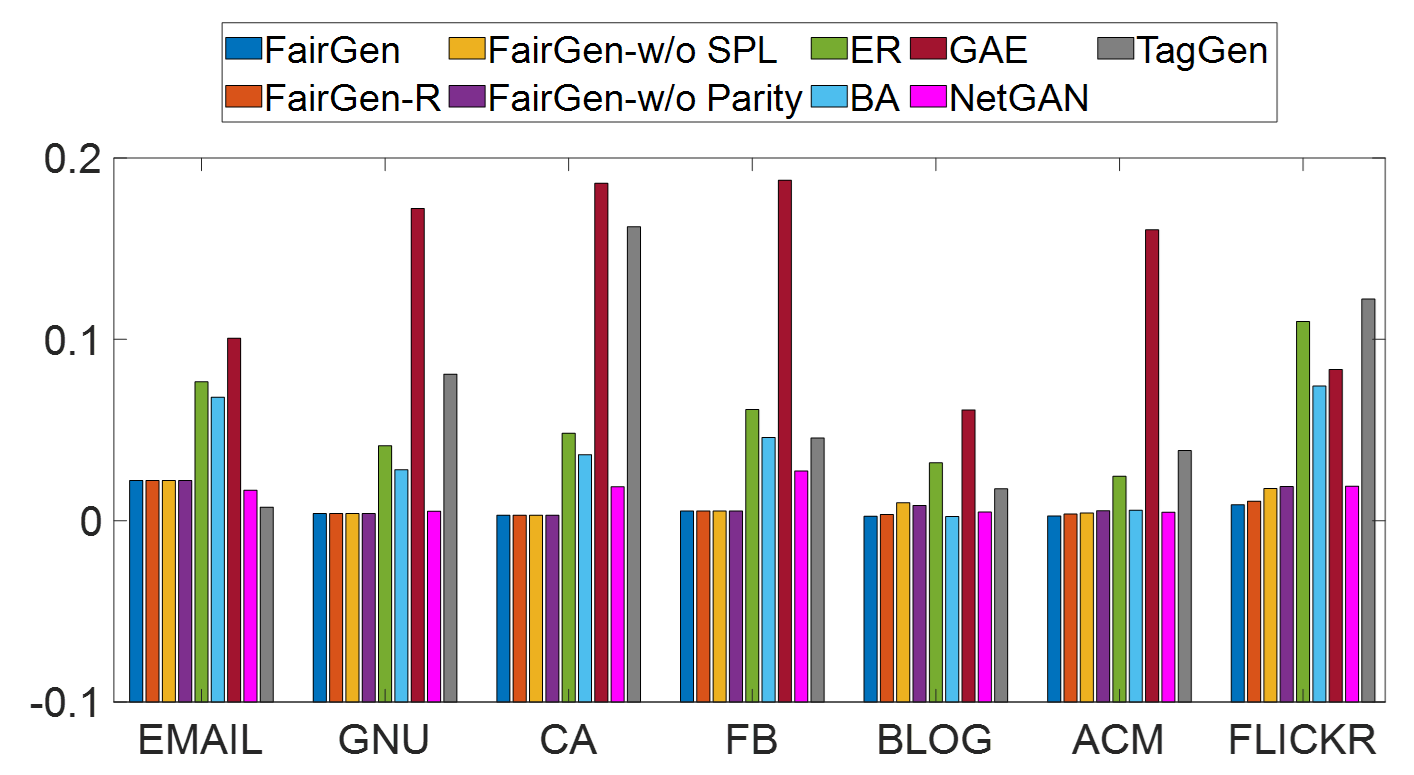}\\
(d) Power Law  Exponent & (e) Gini  &(f) Edge Distribution Entropy\\
\includegraphics[width=0.316\linewidth]{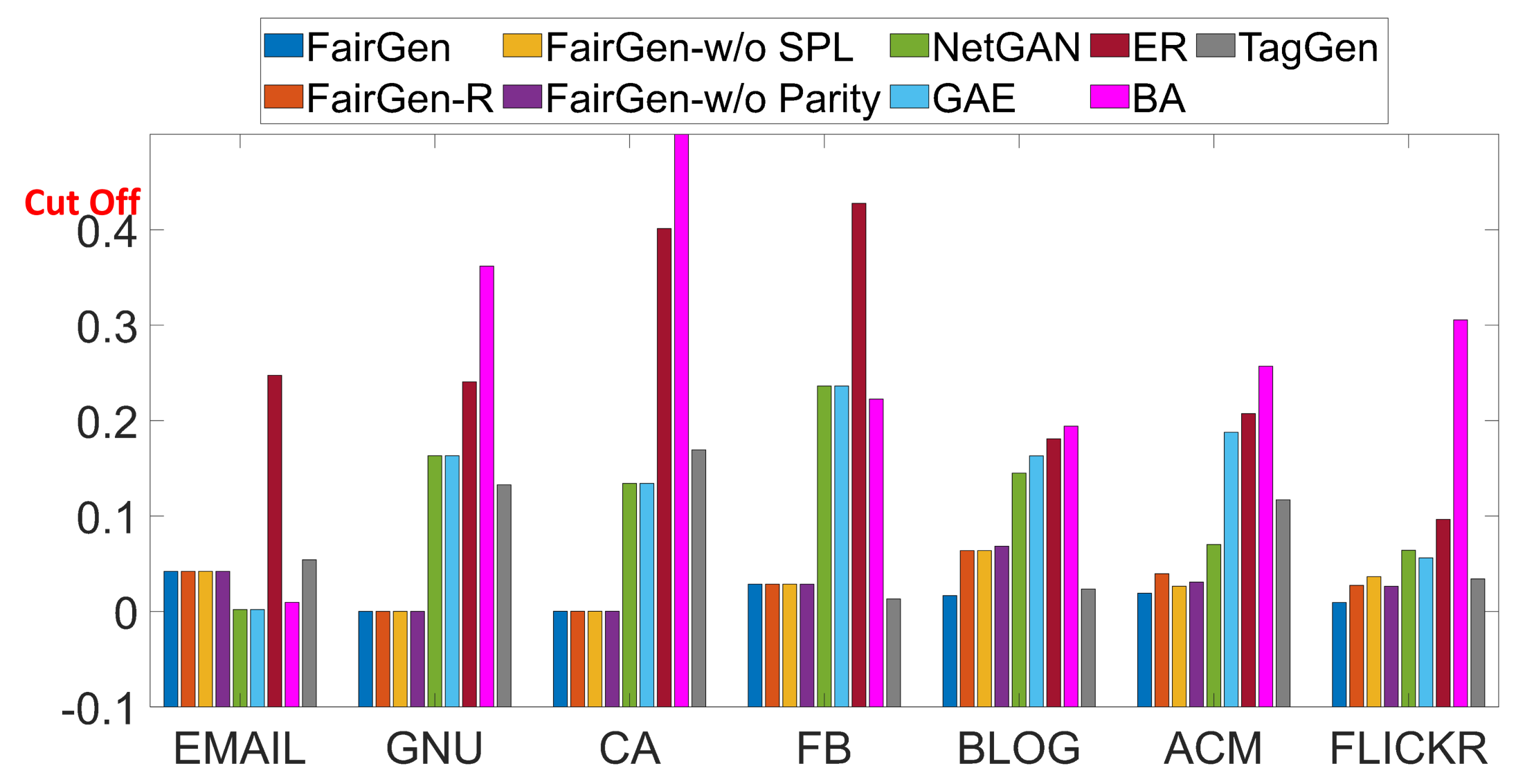}
&\includegraphics[width=0.316\linewidth]{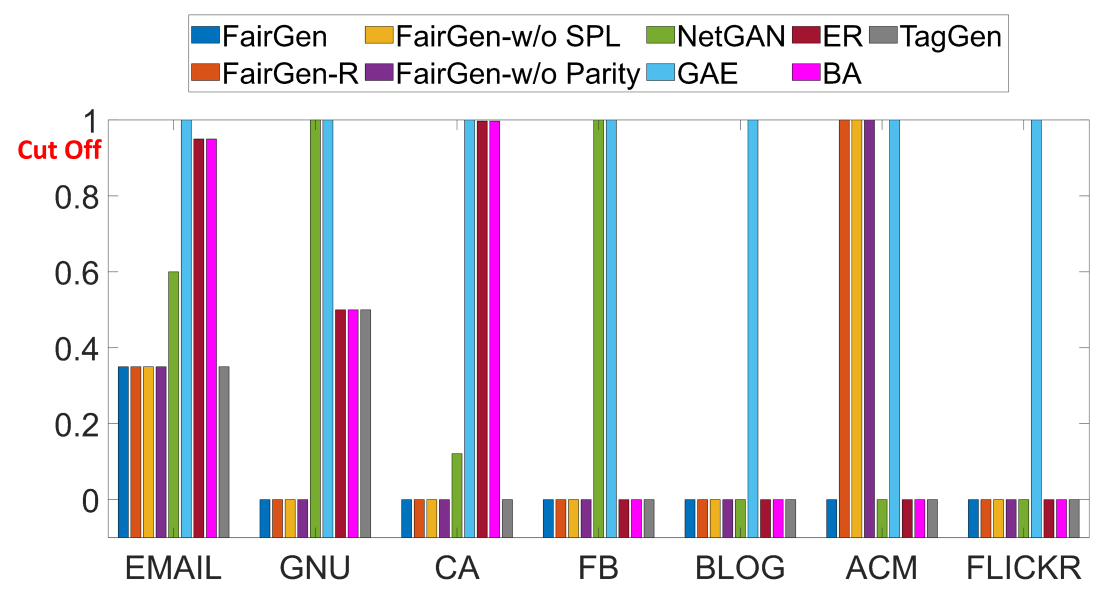}
&\includegraphics[width=0.316\linewidth]{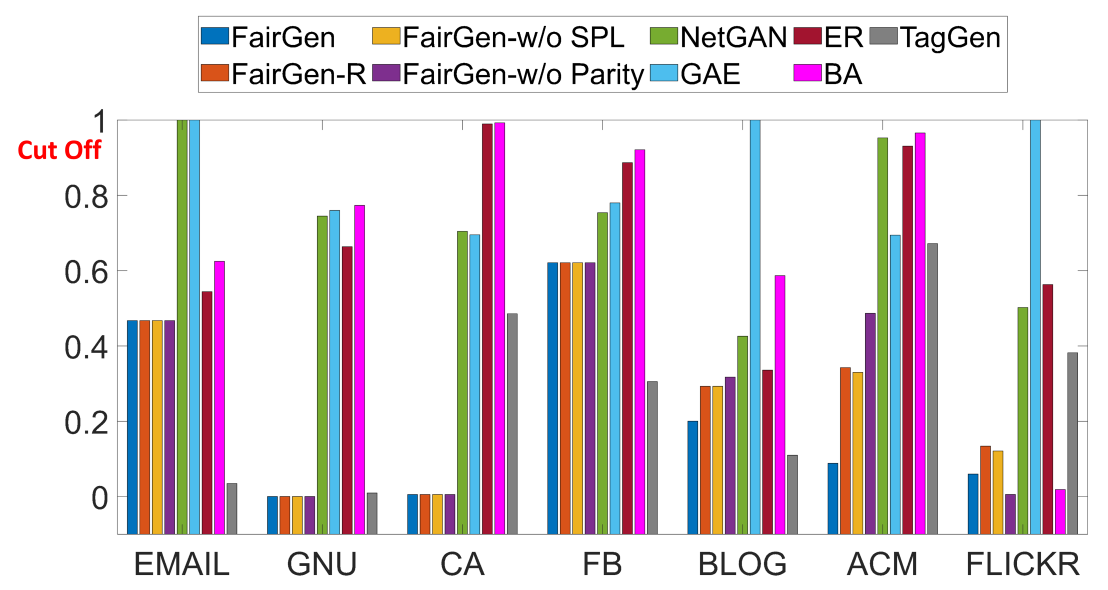}\\
(g) Average Shortest Path Length & (h) Number of Connected Components & (i) Clustering Coefficient \\
\end{tabular}
\end{center}
\caption{Overall discrepancy $R(\mathcal{G}, \widetilde{\mathcal{G}}, f_m)$ regarding nine metrics across seven real networks. We cut off high values for better visibility. The proposed \name\ and its variations (\name-R, \name-w/o SPL, \name-w/o Parity) are the leftmost bars. For better visualization, we change the lower limit of the y-axis for (a), (b), (d), and (f) to -0.1, as the values of some metrics are close to zero. (Smaller metric values indicate better performance)}
\label{fig:no_label}
\end{figure*}

\begin{figure*}[t]
\begin{center}
\begin{tabular}{ccc}
\includegraphics[width=0.316\linewidth]{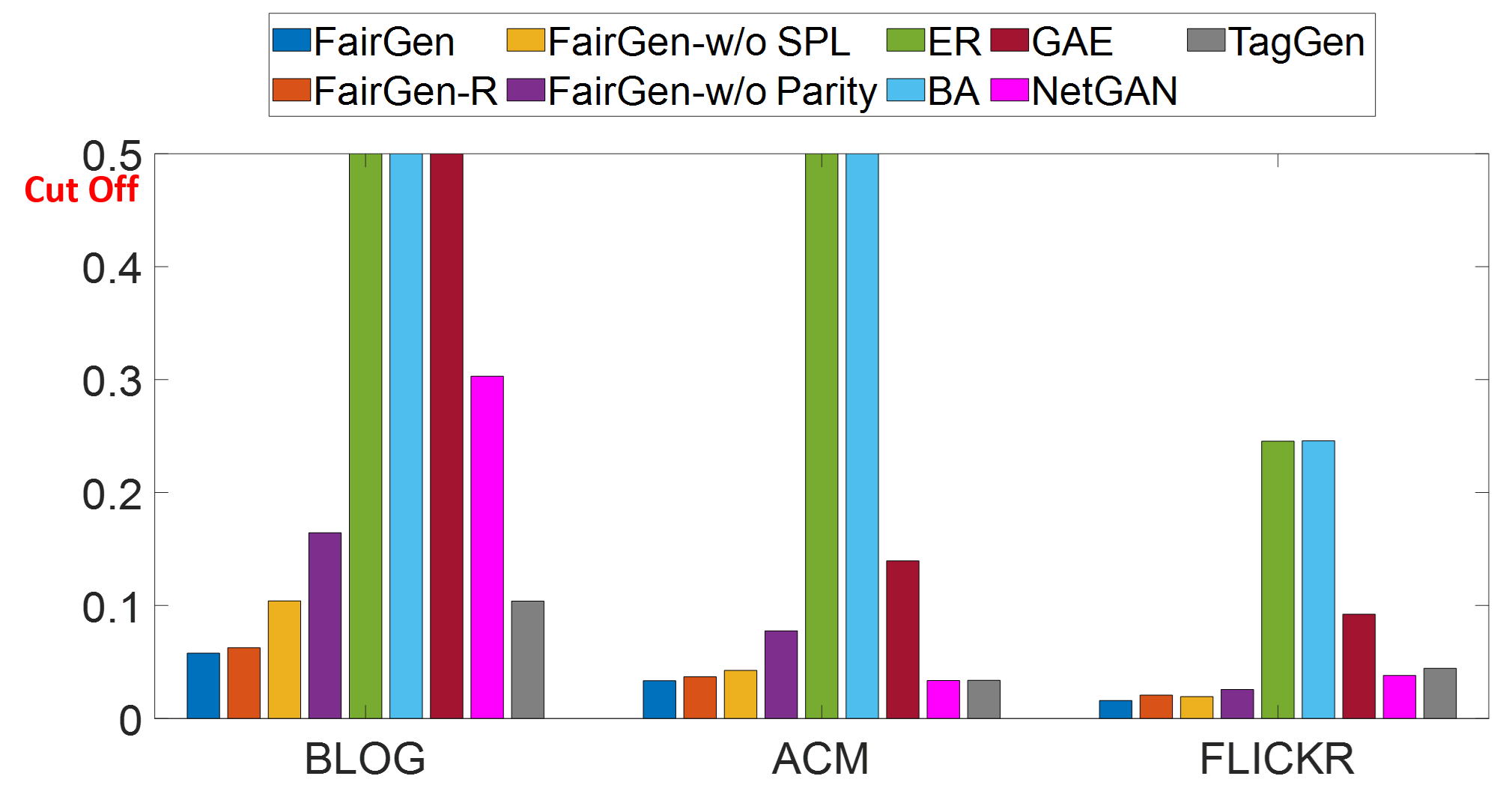} 
&\includegraphics[width=0.316\linewidth]{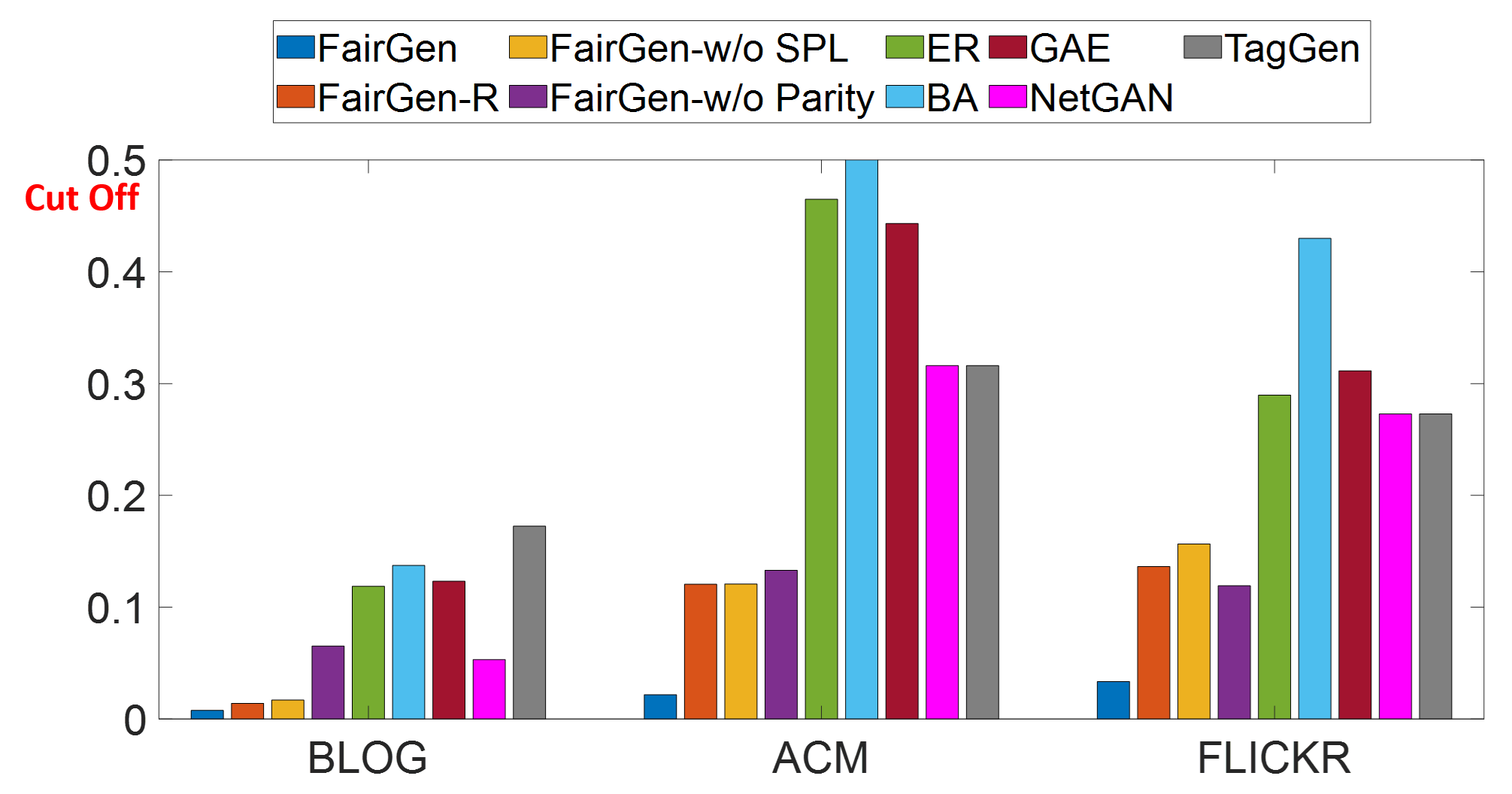}
&\includegraphics[width=0.316\linewidth]{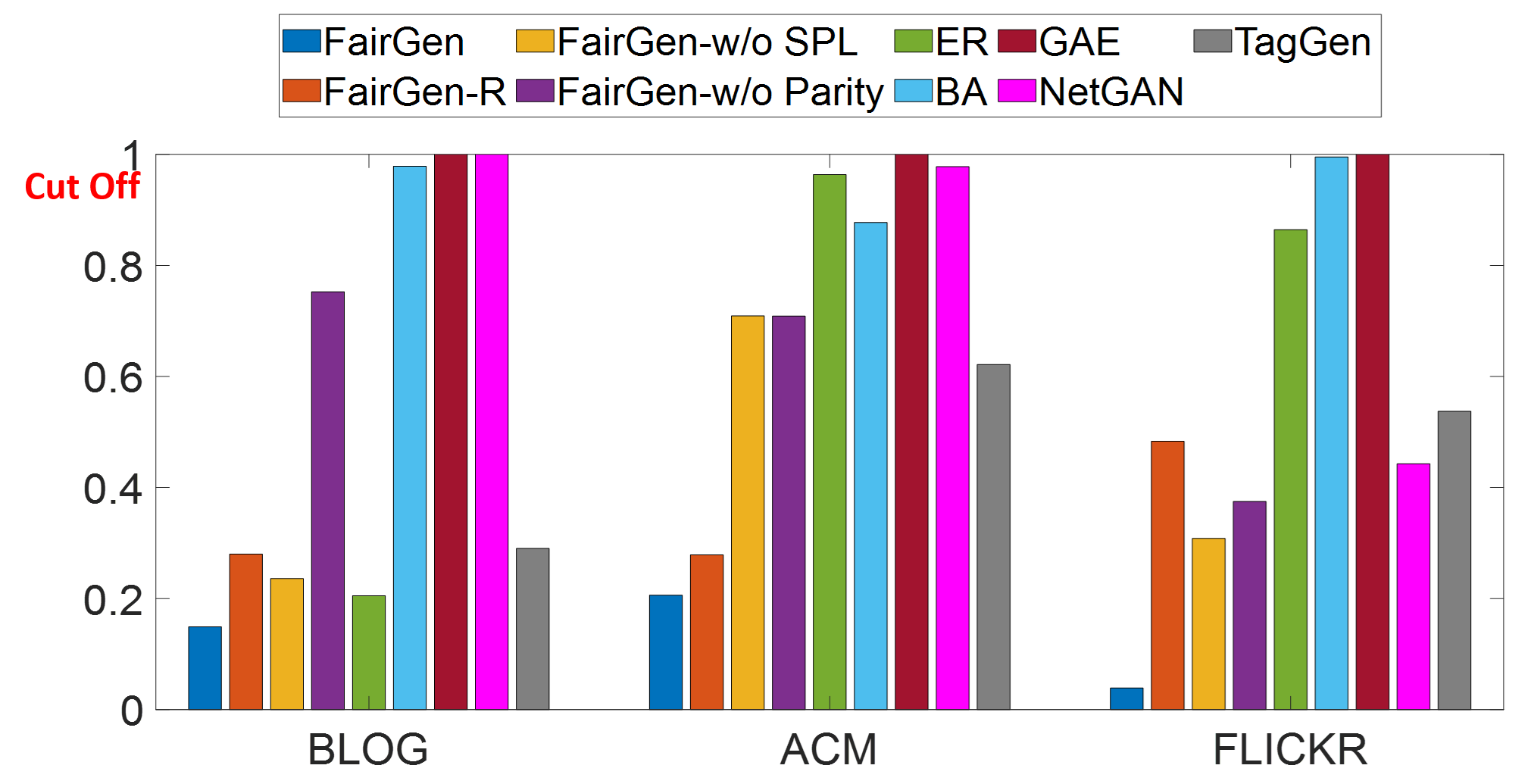} \\
(a) Average Degree  & (b)  LLC & (c) Triangle Count \\
\includegraphics[width=0.316\linewidth]{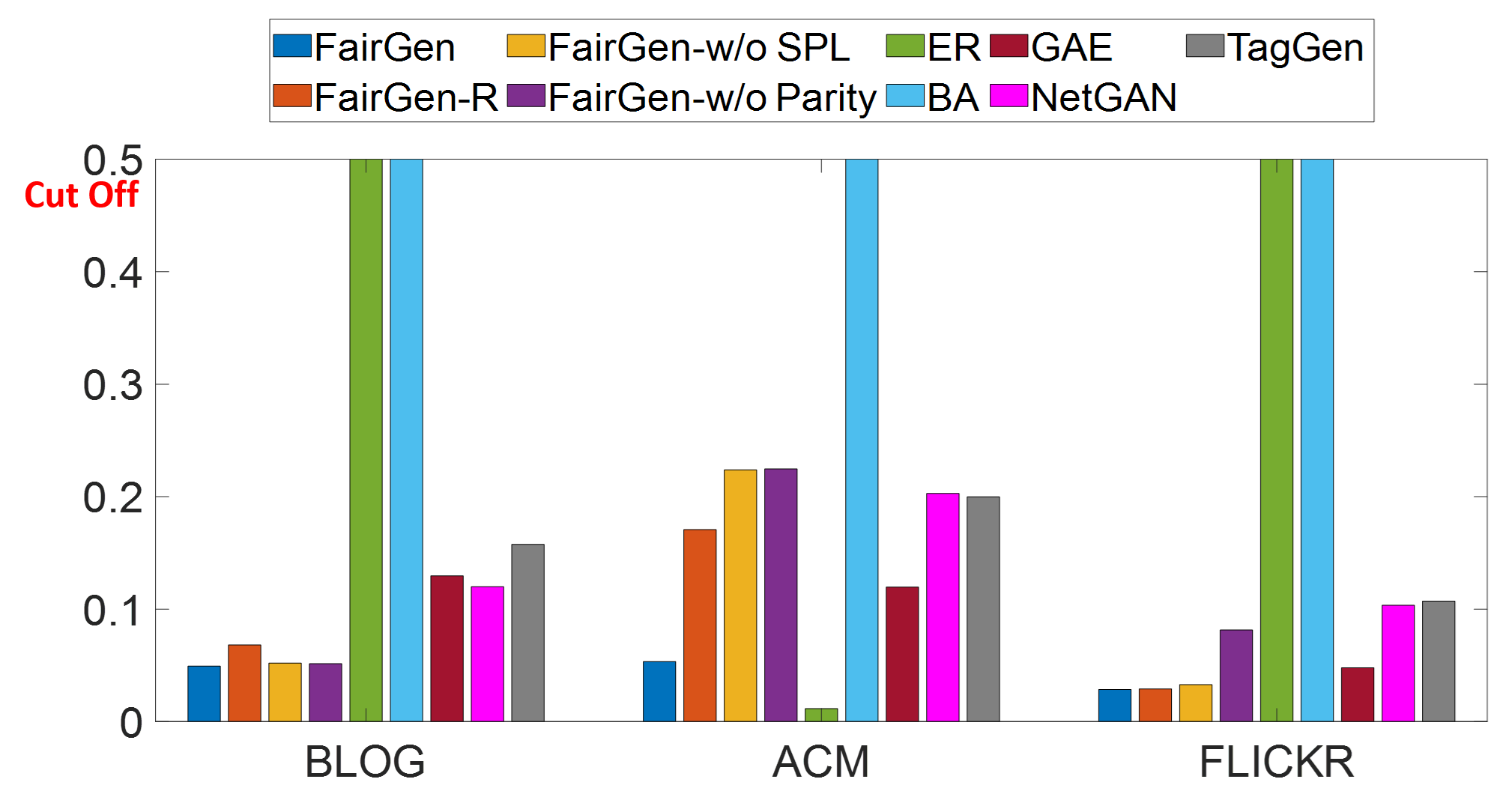}
&\includegraphics[width=0.316\linewidth]{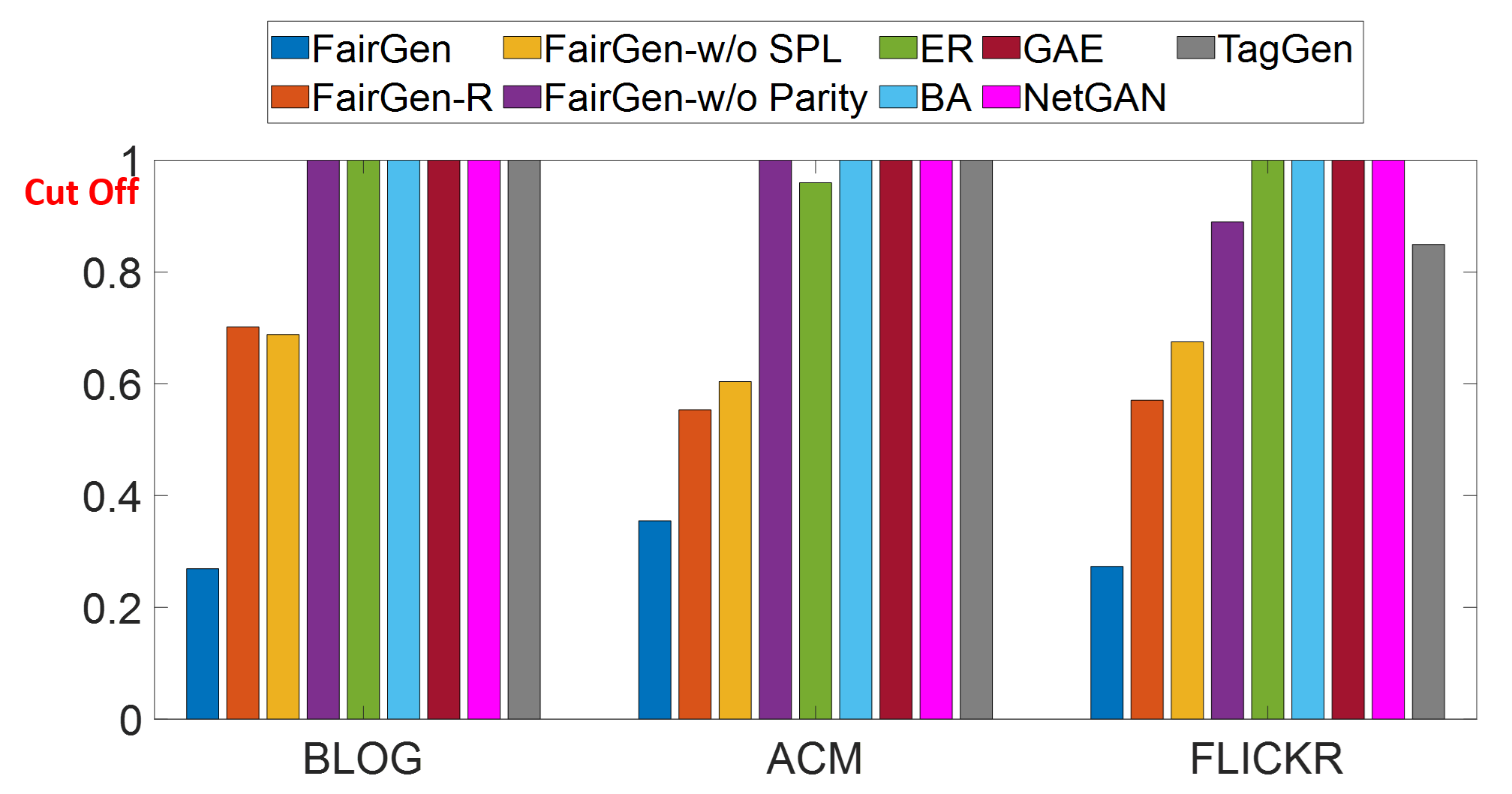}
&\includegraphics[width=0.316\linewidth]{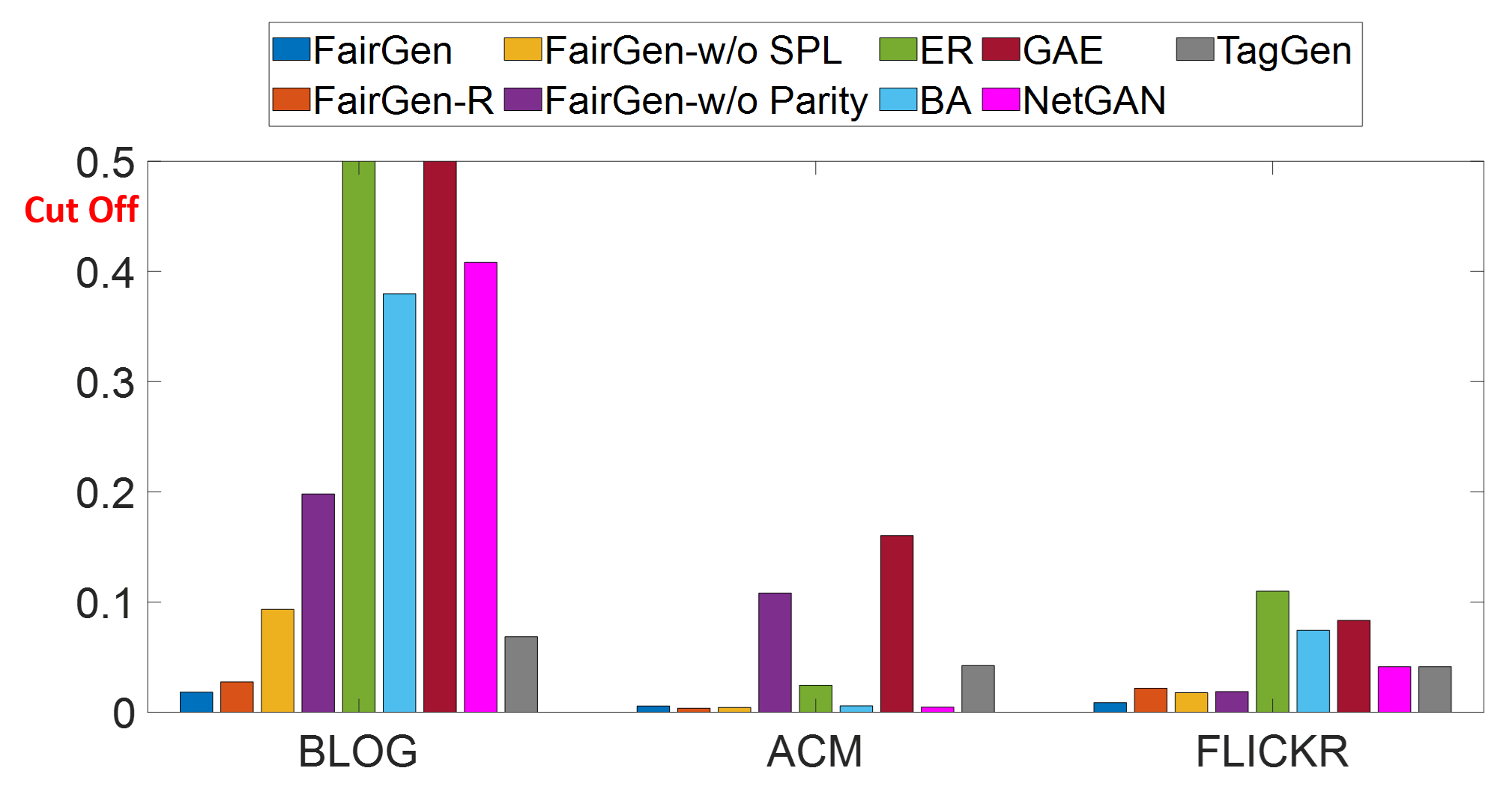}\\
(d) Power Law  Exponent & (e) Gini  &(f) Edge Distribution Entropy\\
\includegraphics[width=0.316\linewidth]{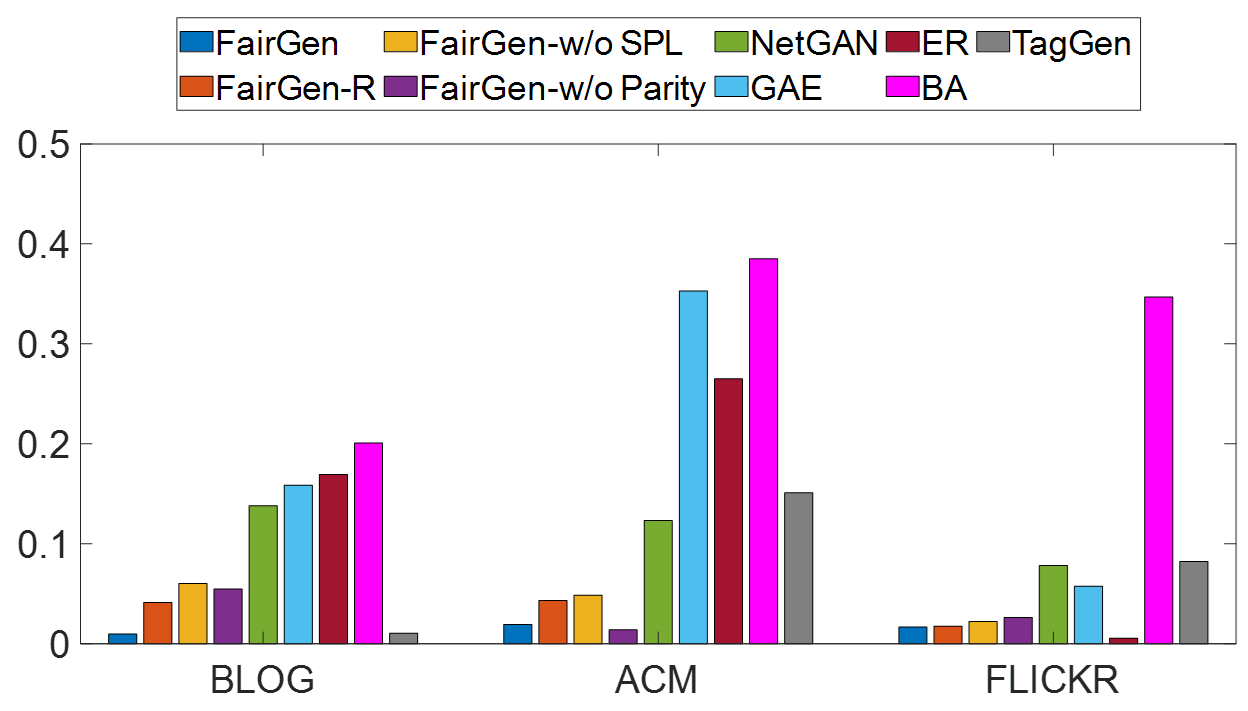}
&\includegraphics[width=0.316\linewidth]{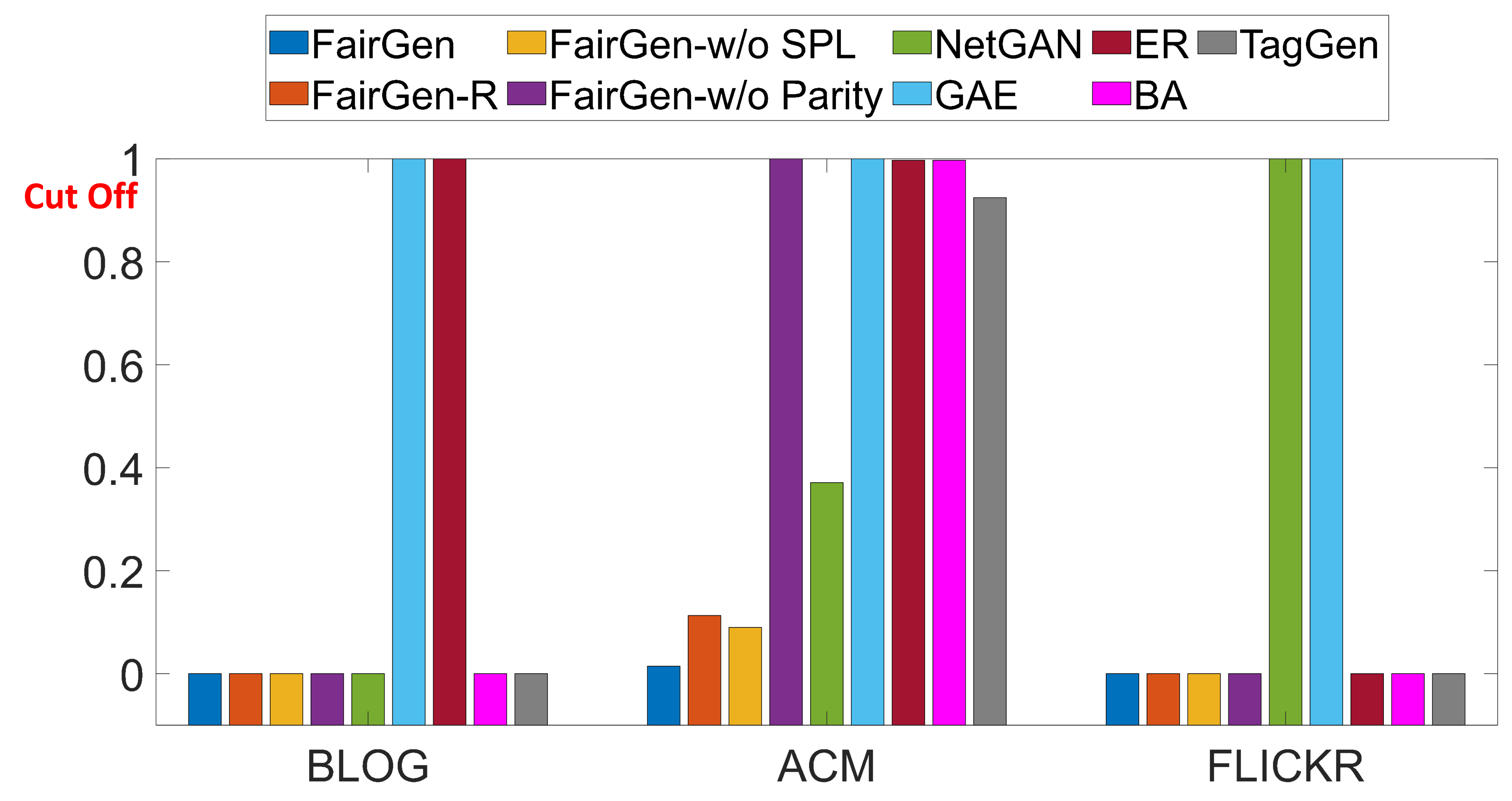}
&\includegraphics[width=0.316\linewidth]{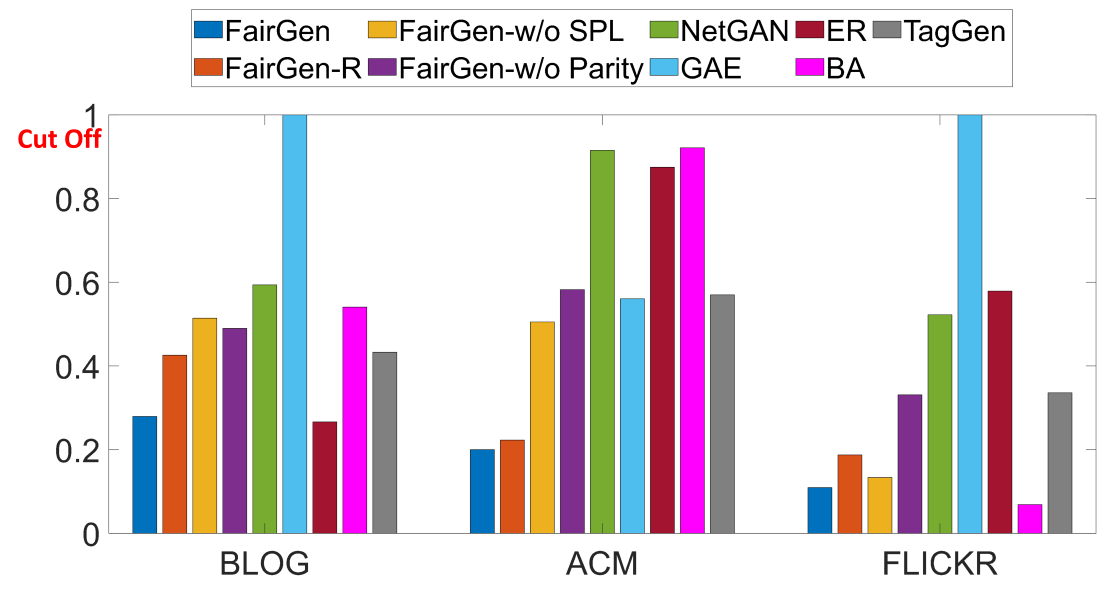}\\
(g) Average Shortest Path Length & (h) Number of Connected Components & (i) Clustering Coefficient \\
\end{tabular}
\end{center}
\caption{Protected group discrepancy $R^+(\mathcal{G}, \widetilde{\mathcal{G}}, \mathcal{S}^+, f_m)$ regarding nine metrics across three real graphs. We cut off high values for better visibility. The proposed \name\ and its variations (\name-R, \name-w/o SPL, \name-w/o Parity) are the leftmost bars. (Smaller metric values indicate better performance)}
\label{fig:label}
\end{figure*}

\subsection{Experiment Setup}
\noindent\textbf{Data Sets:}
We evaluate our proposed algorithm on seven real-world graphs. The statistics of these datasets are summarized in Table~\ref{TB:Net}. Email~\cite{leskovec2015snap} is a student-to-student communication network, where each node represents a student and an edge exists if one student sends one email to another student; FB~\cite{leskovec2015snap} and FLICKR~\cite{tang2009relational} and BLOG~\cite{tang2009relational} are social networks, where each node represents a user and each edge represents one user connected with another user; GNU~\cite{leskovec2015snap} is file-sharing networks, where each node represents a host and each edge indicates the connection between two hosts; CA~\cite{leskovec2015snap} and ACM~\cite{ding2019interactive} are collaboration networks, where each node represents an author and each edge indicates a collaboration between two authors. Particularly, in ACM, BLOG, and FLICKR datasets, the nodes come with the class labels and the memberships of protected group $S^+$ and unprotected group $S^-$. Specifically, the protected group of the FLICKR data set is race; the protected group of the BLOG data set is race; the protected group of the ACM data set is the topic with a small population.

\noindent\textbf{Comparison Methods: } We compare \name\ with multiple graph generative models, including two random graph models, \ie, Erd{\"o}s-R{\'e}nyi (ER) model~\cite{erdos1959random} and Barab{\'a}si-Albert (BA) model~\cite{albert2002statistical}, three deep graph generative models, \ie, GAE~\cite{kipf2016variational}, NetGAN~\cite{bojchevski2018netgan}, TagGen~\cite{DBLP:conf/kdd/ZhouZ0H20}. To investigate the contributions of different parts of \name, we conduct an ablation study by introducing three variations of \name, including \name-R that samples random walks via uniform distribution, \name-w/o-SPL that trains without self-paced learning, and \name-w/o-Parity that trains without the fairness constraint.\\
\textbf{Evaluation: }We present the results regarding the following metrics:
(1) Average Degree (AD): the average node degree; 
(2) LCC: the size of the largest connected component; 
(3) Triangle Count (TC): the count of three mutually connected nodes; 
(4) Power Law  Exponent (PLE): the exponent of the power law distribution of $\mathcal{G}$;
(5) Gini: the Gini coefficient of the degree distribution;  
(6) Edge Distribution Entropy (EDE): the relative edge distribution entropy of $\mathcal{G}$. 
(7) ASPL: Average Shortest Path Length.
(8) NCC: The number of connected components.
(9) CC: Clustering coefficient of a graph.
The formulations of these nine metrics can be found in Table~\ref{table-metric}.
For the sake of easy comparison, we define the overall discrepancy $R(\mathcal{G}, \widetilde{\mathcal{G}}, f_m)$ and the protected set discrepancy $R^+(\mathcal{G}, \widetilde{\mathcal{G}}, \mathcal{S}^+,f_m)$ between the original graph and the generated graph in terms of the above metrics $f_m$.

\begin{equation}\small
    \label{eq: overall discrepancy}
        R(\mathcal{G}, \widetilde{\mathcal{G}}, f_m) = |\frac{f_m(\mathcal{G})- f_m(\widetilde{\mathcal{G}})}{f_m(\mathcal{G})}|
\end{equation}
\begin{equation}\small
    \label{eq: protected group discrepancy}
        R^+(\mathcal{G}, \widetilde{\mathcal{G}}, \mathcal{S}^+,f_m) = |\frac{f_m(\mathcal{G}_{\mathcal{S}^+})- f_m(\widetilde{\mathcal{G}}_{\mathcal{S}^+})}{f_m(\mathcal{G}_{\mathcal{S}^+})}|
\end{equation}
where $\mathcal{G}_{\mathcal{S}^+}$ and $\widetilde{\mathcal{G}}_{\mathcal{S}^+}$ denote the subgraphs that consist of the protected group vertices $\mathcal{S}^+$ in $\mathcal{G}$ and $\widetilde{\mathcal{G}}$, respectively. These subgraphs are the 1-hop ego network with the anchor nodes from the protected group vertices.
Ideally, a fair graph generator should (1) well captures the general structural properties of the input graph $\mathcal{G}$ (small $R(\mathcal{G}, \widetilde{\mathcal{G}}, f_m)$), and also (2) fairly preserves the contextual information of the protected group (small $R^+(\mathcal{G}, \widetilde{\mathcal{G}}, \mathcal{S}^+,f_m)$) in the generated graph $\widetilde{\mathcal{G}}$. 

\begin{table*}
\centering
\caption{Graph statistics for measuring network properties. \lc{}}
\footnotesize
\begin{tabular}{c c l}    
\hline
\textbf{Metric name}    & \textbf{Computation} & \textbf{Description} \\
\hline 
\hline Average Degree (AD)    & $\mathbb{E}[d(v)]$        &  Average node degree.  \\ 
\hline LCC     & $\max_{f \in F} | f|$    &   Size of the largest connected component in $\mathcal{G}$.   \\
\hline Triangle Count (TC)    & $\frac{|\{\{u,v,w\}|\{(u,v),(v,w),(u,w)\}\subseteq \mathcal{E}|}{6}$     &   Number of the triangles. \\ 
\hline Power Law Exponent (PLE)     & $1+n(\sum_{u \in \mathcal{V}} \log(\frac{d(u)}{d_{min}}))^{-1}$    &  Exponent of the power-law distribution of $\mathcal{G}$. \\ 
\hline Gini    & $\frac{2\sum_{i=1}^n i\hat{d}_i}{n \sum_{i=1}^n \hat{d}_i} - \frac{n+1}{n}$    &  Inequality measure for degree distribution.\\
\hline Edge Distribution Entropy (EDE)    & $\frac{1}{\ln n}\sum_{v\in \mathcal{V}} - \frac{d(v)}{|\mathcal{E}|}\ln \frac{d(v)}{|\mathcal{E}|}$   &  Entropy of degree distribution. \\
\hline  ASPL  & $\frac{1}{n(n-1)}\cdot \sum_{i \neq j}d(v_i, v_j)$ & Average Shortest Path Length\\
\hline  NCC & Algorithm 2 in \cite{pearce2005improved} & The number of connected components\\
\hline  CC & $\frac{\text{number of triangles connect to node } v_i}{\text{number of triangles centred around node } v_i}$ & Clustering coefficient of a graph\\
\hline
\end{tabular}
\label{table-metric}
\end{table*}

\subsection{Implementation and Repeatability}
\label{Netfair_metrics}
The experiments are performed on a Windows machine with eight 3.8GHz Intel Cores and a single 16GB RTX 5000 GPU. In our implementation, we set the batch size $N_1 =128$, batch iterations $T_1=3$,  the epoch numbers to be 20, the node embedding dimension to be 100, the number of transformer's heads to be 4, the learning rate to be 0.01, walk length $T=10$, and $\alpha=1$, $\beta=1$, $\gamma=1$.

\subsection{Graph Generation}
We compare the quality of the generated graphs with eight baseline methods at the level of both the entire graph $\widetilde{\mathcal{G}}$ and the protected group $\mathcal{S}^+$ in terms of nine classic graph properties. 
We fit all the models on the seven real-world graphs and report the statistics of the generated graphs in Figure~\ref{fig:no_label} and Figure~~\ref{fig:label}. 
In Figure~\ref{fig:no_label}, we provide the comparison results in terms of the overall discrepancy $R(\mathcal{G}, \widetilde{\mathcal{G}}, f_m)$ and have the following observations. (1) The traditional random graph models (\ie, ER, BA) excel at recovering the corresponding structural properties (\eg, Largest Connected components, Poisson degree distribution and heavy-tailed degree distribution) that they aim to model, whereas they fail to deal with the ones (\eg, triangle count) that they do not account for. (2) The deep graph generative models (\eg, \name, NetGAN, TagGen) have better generalization to different network properties than the random graph models. (3) NetGAN performs better than \name\ on the data sets that provide labels and the protected group information, such as the FLICKR data set in Figure~\ref{fig:no_label} (c). This is consistent with the objective of \name, which is not merely minimizing the overall reconstruction loss of the observed graphs. (4) \name\ achieves comparable and even better performance than the baseline methods in most cases. By incorporating the label information and fairness constraint to protect the protected group nodes, \name\ slightly sacrifices the overall discrepancy to some extent. Notice that based on the statistics shown in Table~\ref{TB:Net}, the ratio of the protected group is significantly smaller than the unprotected group on BLOG, ACM and FLICKR graphs, and generated graphs tend to be biased if we only evaluate the performance of the overall discrepancy. Thus, we further measure the discrepancy of the protected group for all methods in Figure~\ref{fig:label}. In particular, we observe that \name\ consistently outperforms all the other methods across three data sets on all nine metrics in terms of the protected group discrepancy, which demonstrates how well the protected group is preserved in the generated graphs.

In addition, we conduct the ablation study to examine the effectiveness of the proposed sampling strategy $f_S(\cdot)$. The experimental results are shown in Table~\ref{ablation_study}, where `Negative Sampling' refers to the variant of \name\ by replacing $f_S(\cdot)$ with the negative sampling strategy used in Node2vec. A smaller value indicates better performance. By observation, we find that \name\ achieves the smallest discrepancy of $R^+(\mathcal{G}, \widetilde{\mathcal{G}}, \mathcal{S}^+,f_m)$ comparing it with the negative sampling strategy. This observation suggests that the proposed sampling strategy is better than negative sampling.

\begin{table*}
\centering
\caption{Ablation study regarding different sampling strategies with respect to $R^+(\mathcal{G}, \widetilde{\mathcal{G}}, \mathcal{S}^+,f_m)$. Negative sampling refers to the variant of \name\ by replacing $f_S(\cdot)$ with negative sampling. A smaller value indicates better performance.}
\footnotesize
\begin{tabular}{|c|c|c|c|c|c|c|c|c|c|}    
\hline Method (Dataset)             & AD      &  LLC     & TC       & PLE      & Gini     & EDE      & ASPL     & NCC      & CC\\ 
\hline Negative Sampling (BLOG)     & 0.0625  &  0.0140  &  0.2801  &  0.0682  &  0.7016  &  0.0276  &  0.0000  &  0.4257  &  0.0413  \\ 
\hline \name\ (BLOG)                  & 0.0577  &  0.0077  &  0.1492  &  0.0493  &  0.2692  &  0.0182  &  0.0000  &  0.2793  &  0.0096  \\ 
\hline Negative Sampling (ACM)      & 0.0369  &  0.1205  &  0.2787  &  0.1707  &  0.5535  &  0.0036  &  0.1130  &  0.2229  &  0.0432  \\ 
\hline \name\ (ACM)                 & 0.0334  &  0.0218  &  0.2062  &  0.0534  &  0.3549  &  0.0056  &  0.0145  &  0.2001  &  0.0193  \\ 
\hline Negative Sampling (FLICKR)   & 0.0206  &  0.1364  &  0.4834  &  0.0291  &  0.5705  &  0.0218  &  0.0000  &  0.1874  &  0.0175  \\ 
\hline \name\ (FLICKR)               & 0.0158  &  0.0335  &  0.0389  &  0.0285  &  0.2732  &  0.0088  &  0.0000  &  0.1094  &  0.0166  \\ 
\hline
\end{tabular}
\label{ablation_study}
\end{table*}

\subsection{Data Augmentation}
\begin{figure}[htp]
\centering  
\includegraphics[width=0.85\linewidth]{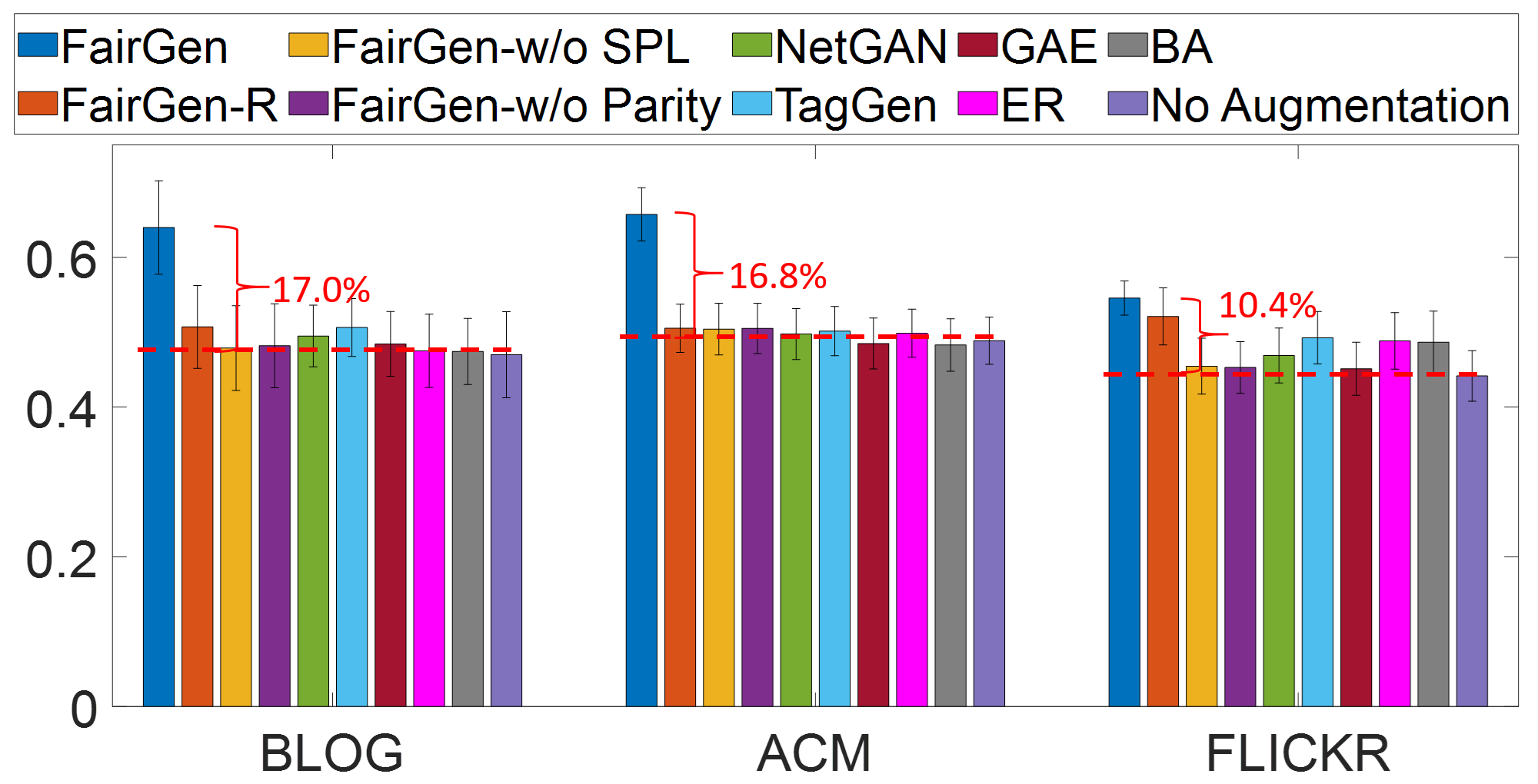}\\
\caption{Data augmentation for node classification. The red dotted line shows the performance without data augmentation. \name\ and its variations are the leftmost bars. (Larger metric values indicate better performance)}
\label{fig_augmentation}
\end{figure}

Figure~\ref{fig:no_label} and Figure~\ref{fig:label} show the effectiveness of our proposed method and how well the generated graph by \name~ preserves the structural information with fairness constraint.
Next, we conduct a case study to evaluate the capability of \name\ further in augmenting the performance of a prediction model for node classification. Specifically, we aim to see how well our label-informed generative model boosts the performance of the node classification task via data augmentation by comparing it with the one without data augmentation.
Here are the procedures of the case study. First, we employ a logistic regression classifier as our base model, which is trained on the learned graph embedding of the original graph via node2vec~\cite{grover2016node2vec}. Then, we aim to produce potential edges for the original graph by a specific graph generative model and insert $5\%$ more edges into the original graph to augment the data. Next, we retrain the node2vec on the augmented graphs and use the learned logistic regression model to predict the label. Notice that 'No Augmentation' in Figure~\ref{fig_augmentation} refers to the node classification task performed on the original graph without any augmentation.
In our experiments, we split the data set into ten folds, with 90\% for training and 10\% for testing. In Figure~\ref{fig_augmentation}, we provide the accuracy score (\ie, bar height) as well as the standard deviation (\ie, error bars) in the task of node classification on BLOG, ACM, and FLICKR data sets. In general, we observed that: (1) \name\ significantly outperforms all the other graph generative models regarding performance improvement; (2) the baseline methods (\eg, GAE, NetGAN, TagGen, etc.) without utilizing label information can only marginally increase the performance. For example, on the BLOG graph, \name\ boosts the performance to $17\%$, while the best competitor \name-R and the second-best competitor TagGen only achieve $3.7\%$ and $3.6\%$ improvement in accuracy over the performance of no augmentation, respectively.

\subsection{Parameter Sensitivity Analysis}
\begin{figure}[t]
\centering 
\begin{tabular}{cc}
\includegraphics[width=0.45\linewidth]{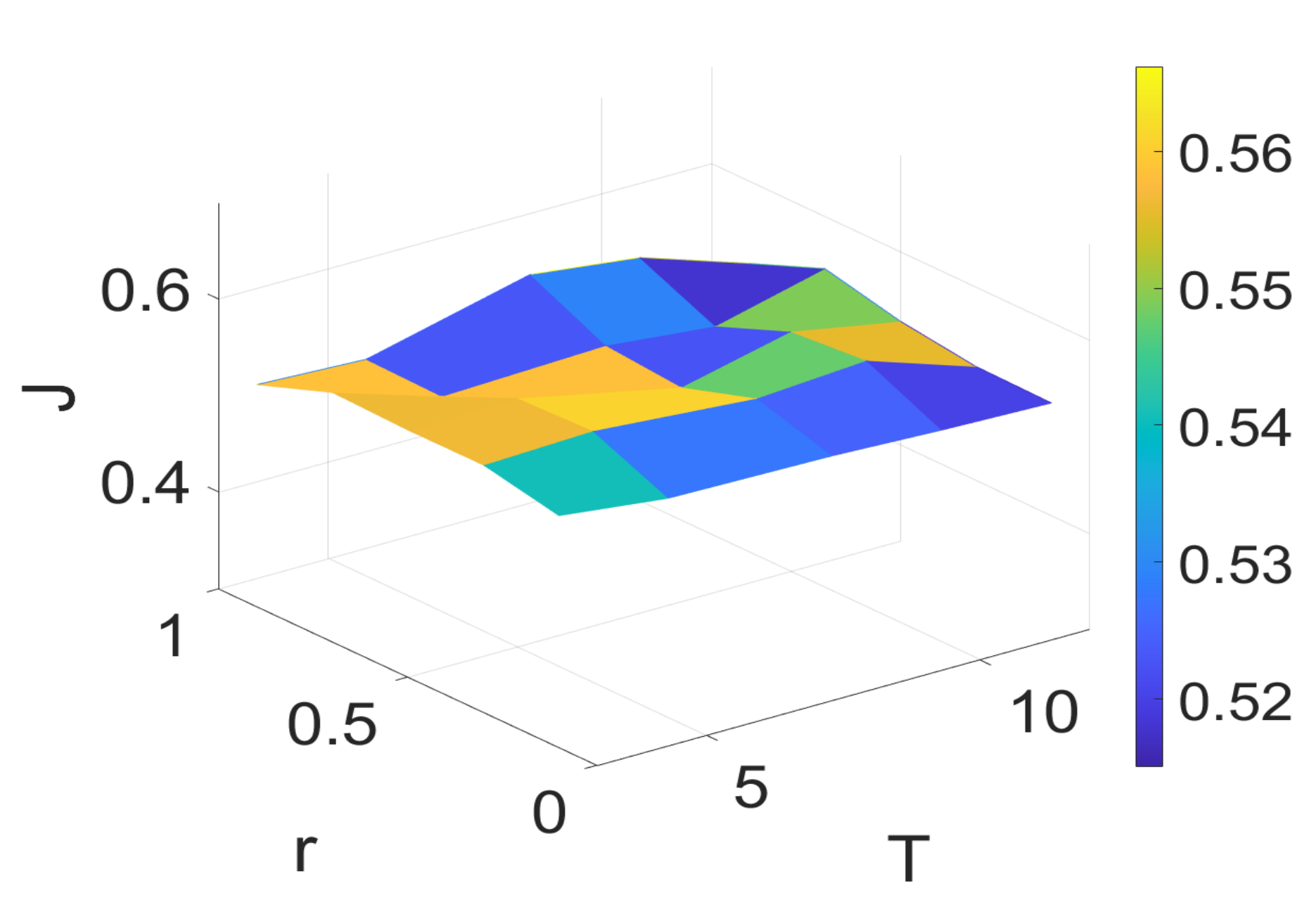}
&\includegraphics[width=0.45\linewidth]{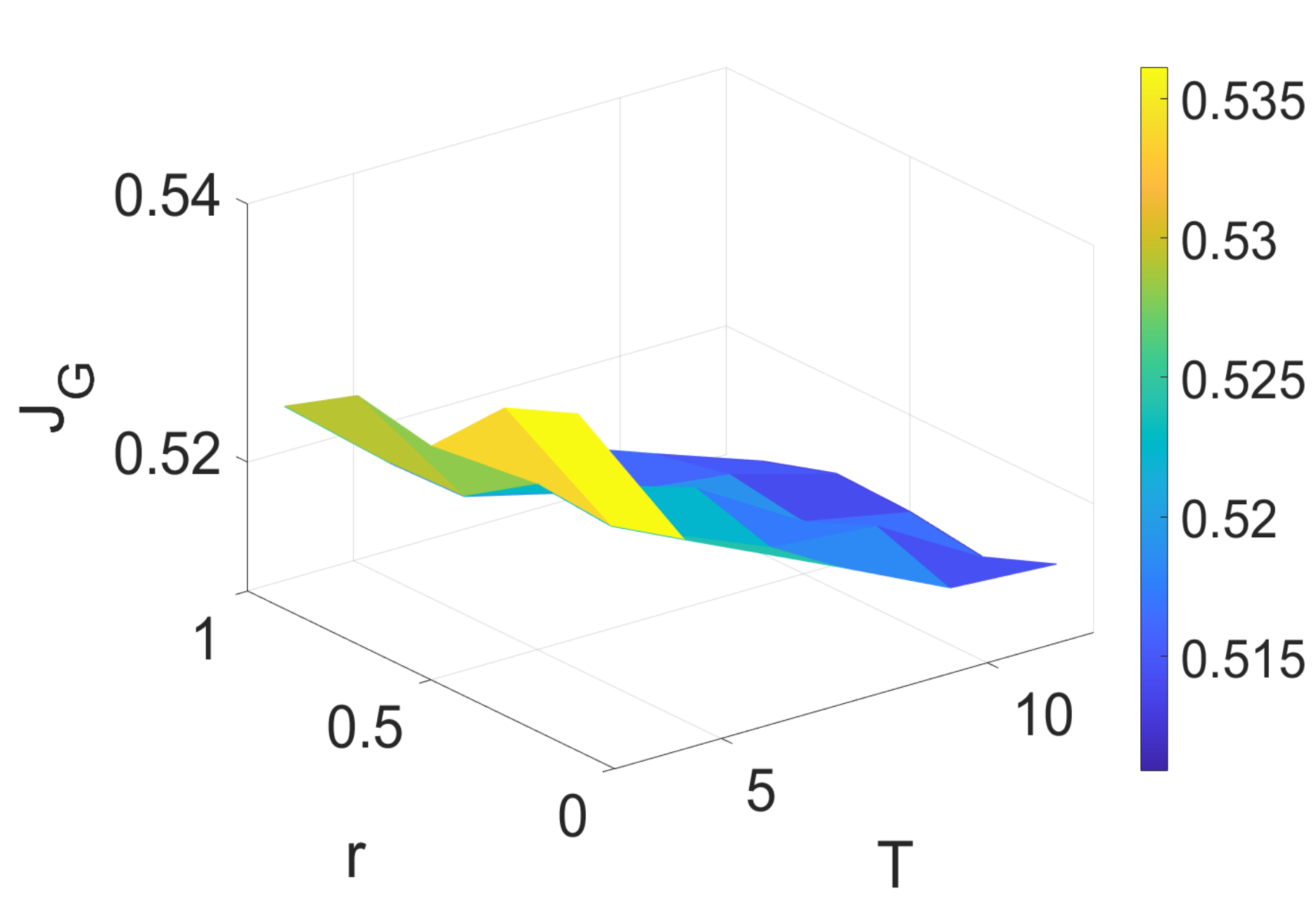}\\
\makecell[c]{(a) }
&\makecell[c]{(b) }\\
\includegraphics[width=0.45\linewidth]{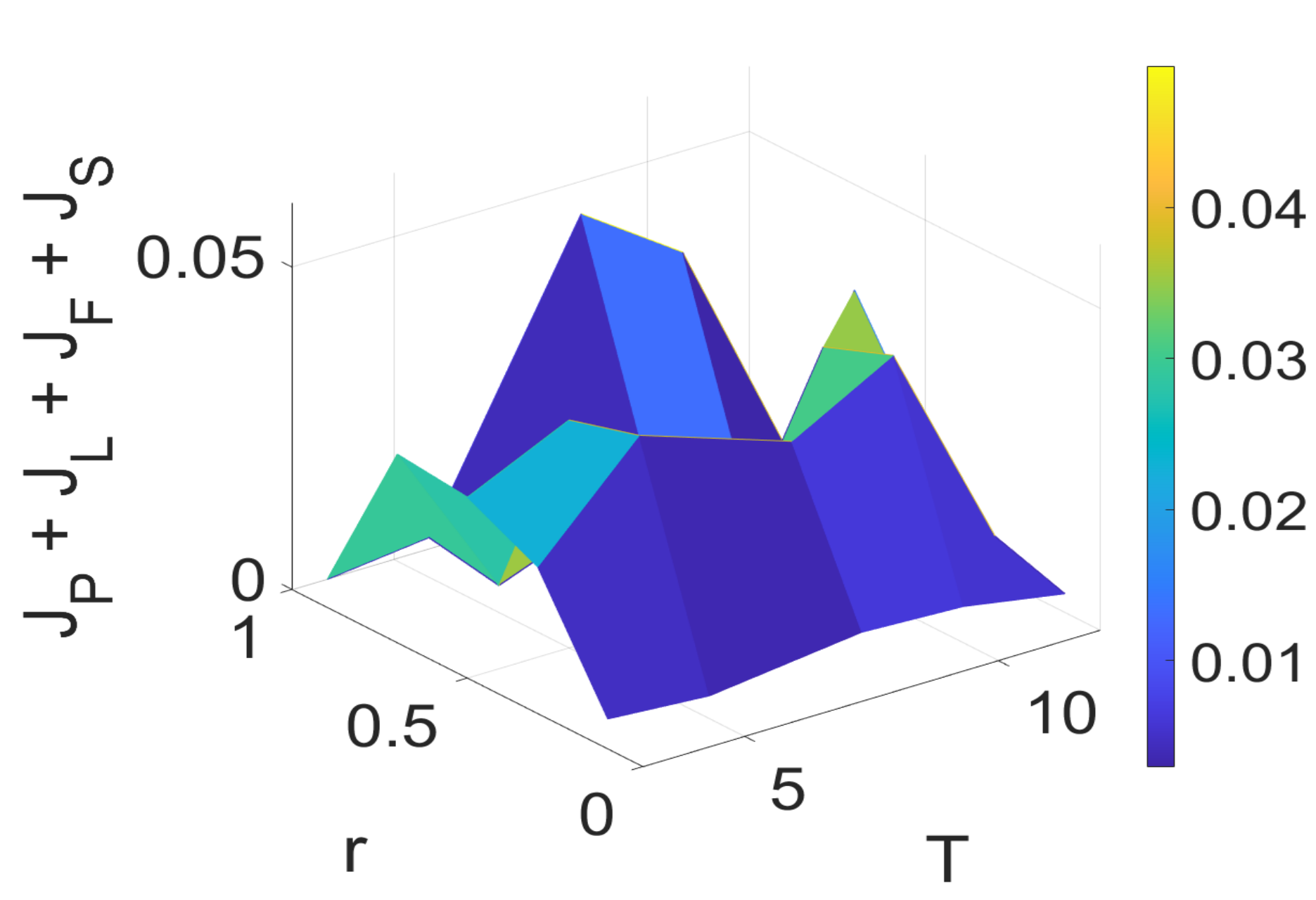}
&\includegraphics[width=0.45\linewidth]{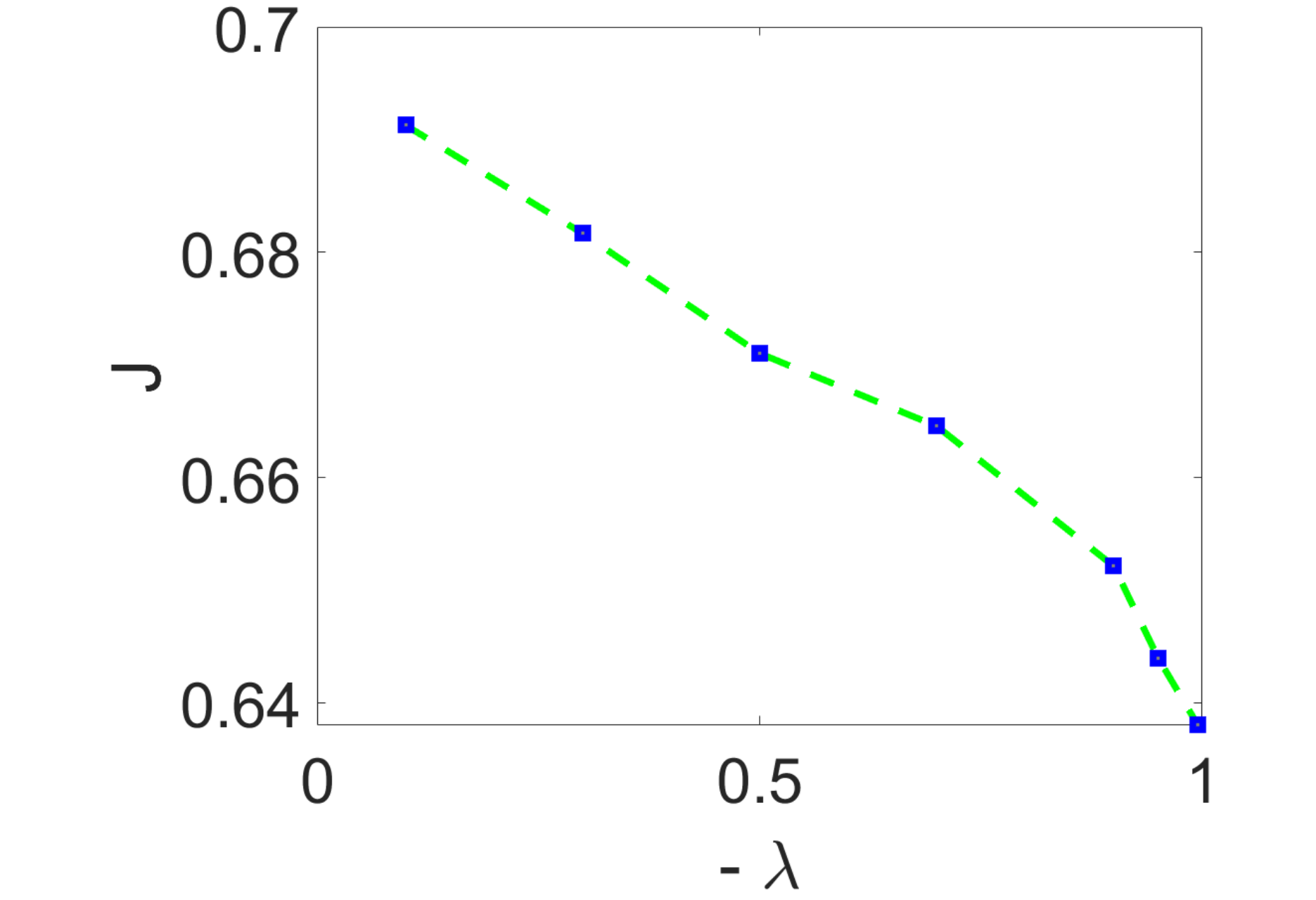}\\
\makecell[c]{(c) }
&\makecell[c]{(d) }\\
\end{tabular}
\caption{Parameter sensitivity analysis. 
(a) Overall loss w.r.t. walking length $T$ \& sampling ratio $r$; (b) Generator loss w.r.t.  walking length $T$ \& sampling ratio $r$; (c) Discriminator loss w.r.t. walking length $T$ \& sampling ratio $r$; (d) Overall loss w.r.t. learning threshold $- \lambda$.
}
\label{fig: context sampling parameters}

\end{figure}

We investigate the sensitivity of both context sampling parameters (\ie, walking length $T$ and sampling ratio $r$) and also self-paced learning parameters (\ie,  learning threshold $\lambda$). In Figure~\ref{fig: context sampling parameters}(a)(b)(c), we individually report the overall loss $\mathcal{J}$, generator loss $\mathcal{J}_G$, and discriminator loss $\mathcal{J}_P + \mathcal{J}_L + \mathcal{J}_F + \mathcal{J}_S$ for different settings of walking length $T$ and sampling ratio $r\in [0,1]$. We observe that 
(1) the overall loss  $\mathcal{J}$ is generally smooth across different settings of $T$ and $r$; (2) a major component of the overall loss $\mathcal{J}$ comes from the generator, which is largely consistent with our intuition. As the output space of the generator (\ie, the entire graph with $O(n^2)$ space complexity) is much larger than the output space of the discriminator (\ie, node labels with $O(n)$ space complexity), the learning complexity of the generator is significantly higher than the one of the discriminator. Moreover, in Figure~\ref{fig: context sampling parameters}(c), the discriminator obtains the largest loss when $r$ is around 0.5 while getting the lowest loss when $r$ is close to 0 or 1. This is because when $r=1$, our context sampling strategy $f_S(\cdot)$ purely extracts the general network context information without label guidance; when $r=0$, our context sampling strategy $f_S(\cdot)$ entirely extracts network context with the guidance of labels obtained from the discriminator at the last iteration of self-paced learning; when $r = 0.5$,  $f_S(\cdot)$ extracts both general network context and label-informed network context. That is to say, when $r = 0.5$, \name\ blends the two kinds of network context information equally into a unique embedding space, which leads to a higher discriminator loss. 
In Figure~\ref{fig: context sampling parameters}(d), we present the overall loss with respect to different settings of the learning threshold $ - \lambda$. Intuitively, when $ -\lambda$ is large, \name\ only propagates labels to the unlabeled nodes with a high confidence score $\log Pr(\hat{y_i} = c| {x_i}) >  - \lambda$ for training in the next iteration. Thus, we can see $J$ obtains a lower loss when $-\lambda$ is close to 1 but obtains a higher loss when $-\lambda$ is around $0$.

\subsection{Scalability Analysis }
\label{scalability_analysis}
\begin{figure}[htp]
\centering 
\begin{tabular}{cc}
\hspace{-0.3cm}
\includegraphics[width=0.47\linewidth]{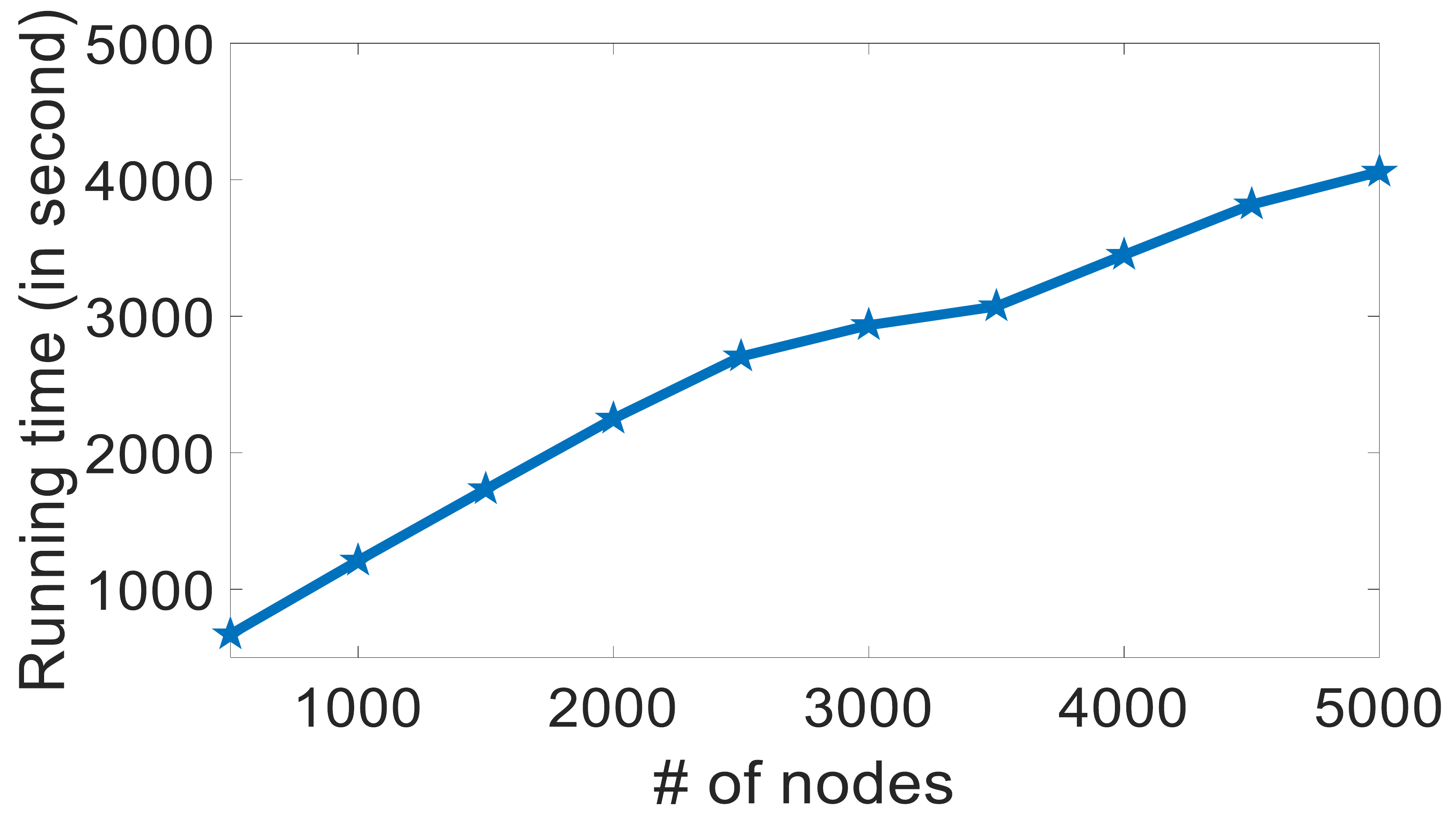} &\includegraphics[width=0.47\linewidth]{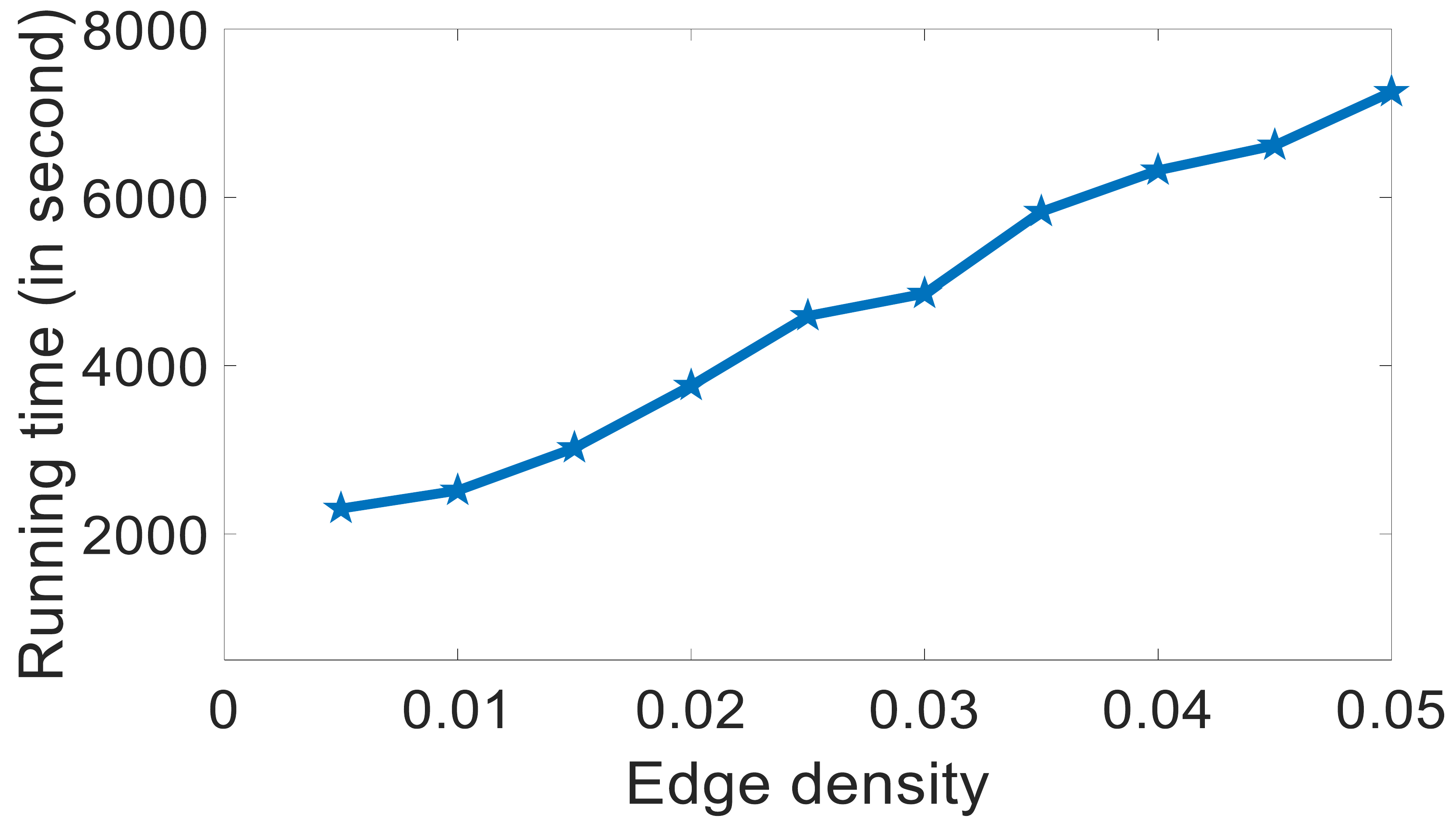}\\
(a) Running time v.s  &(b) Running time v.s \\
\# of nodes & edge density\\
\end{tabular}
\caption{Scalability analysis.}
\label{fig_analysis_1}
\vspace{-0.5cm}
\end{figure}

Here, we analyze the scalability of \name, by reporting the running time of \name\ on a series of synthetic graphs with increasing sizes (\ie, the number of nodes and the edge density). To control the number of nodes and the edge density, we generate the synthetic graphs via ER algorithm~\cite{erdos1959random}. 
In Figure~\ref{fig_analysis_1} (a), we fix the edge density to be 0.005 and gradually increase the number of nodes from 500 to 5000. In Figure~\ref{fig_analysis_1} (b), we fix the number of nodes to 5,000 and increase the edge density from 0.005 to 0.05. Based on the results in Figure~\ref{fig_analysis_1}, we observe that the complexity of the proposed method is almost linear to both the number of nodes and the edge density, which is desirable for modeling large-scale networks. 

Furthermore, we also report the running time for different baseline methods on seven benchmark datasets shown in Table~\ref{running_time}. Notice that ER and BA do not have a training phase but a generation phase. Thus, the running time of these two methods is much less than deep learning-based methods, such as GAE, NETGAN, TagGen, and \name. By observation, the running time of our proposed method is much less than NetGAN but it achieves better performance than NetGAN as shown in Figures~\ref{fig:no_label} and \ref{fig:label}.

\begin{table*}
\centering
\caption{Running time (in seconds) of different baseline methods }
\footnotesize 
\begin{tabular}{|c|c|c|c|c|c|c|c|}    
\hline
\textbf{Method}    & \textbf{EMAIL} & \textbf{GNU} & \textbf{CA} & \textbf{FB} & \textbf{BLOG} & \textbf{ACM} & \textbf{FLICKR}\\
\hline ER           & 0.093     &  0.109    &  0.078    &  0.469    &  0.938    &  1.860    &  1.423    \\ 
\hline BA           & 0.015     &  0.140    &  0.094    &  0.094    &  0.293    &  1.374    &  0.841    \\ 
\hline GAE          & 57.12     &  372.31   &  258.68   &  422.18   &  741.02   &  2889.07  &  1339.1   \\ 
\hline NetGAN       & 1397.36   &  8323.7   &  5643.21  &  3218.64  &  6036.42  &  29688.28 &  7834.12  \\ 
\hline TagGen       & 372.63    &  2162.13  &  1707.01  &  971.23   &  1462.37  &  7253.16  &  1632.76  \\ 
\hline FairGen      & 394.65    &  2254.37  &  1768.25  &  1013.66  &  3248.86  &  11429.91 &  4969.56  \\ 
\hline
\end{tabular}
\label{running_time}
\end{table*}

\begin{figure*}
\begin{center}
\begin{tabular}{ccccc}
\includegraphics[width=0.18\linewidth]{fig/original_graph_no_edge.png}
&\includegraphics[width=0.18\linewidth]{fig/netgan_2000_no_edge.png}
&\includegraphics[width=0.18\linewidth]{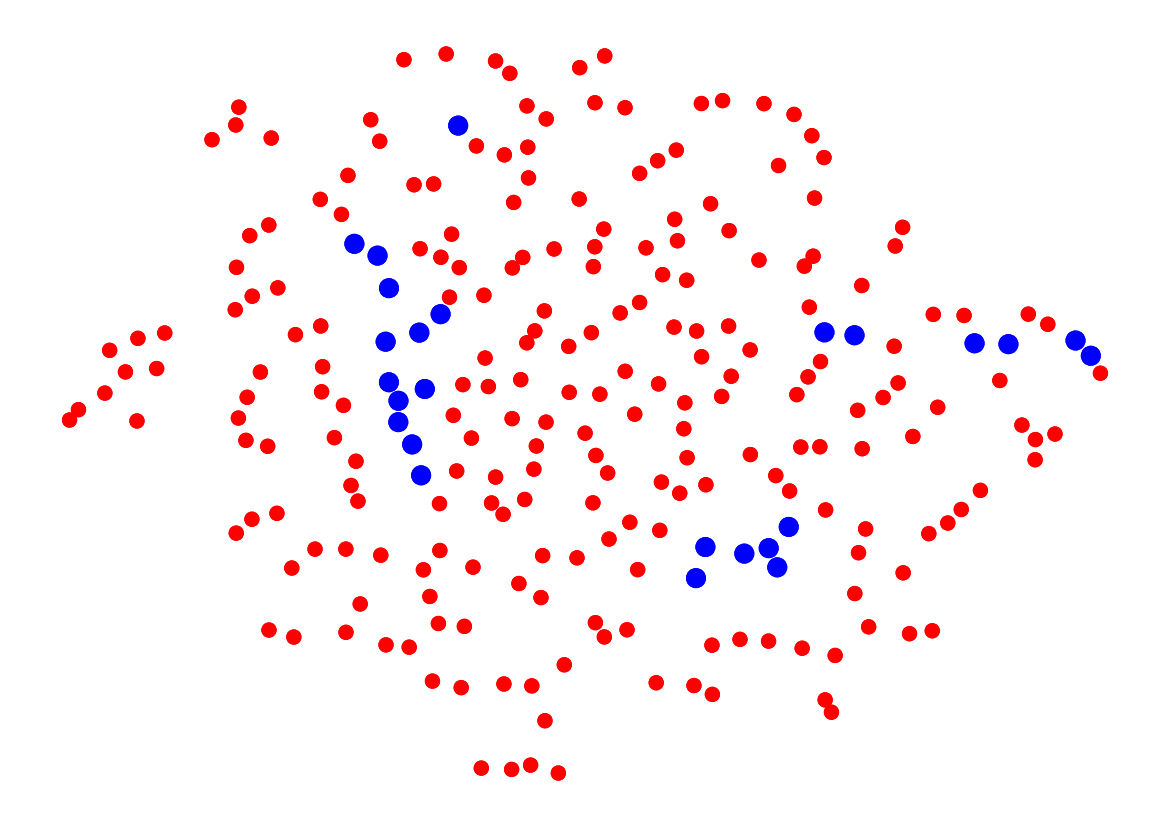} 
&\includegraphics[width=0.18\linewidth]{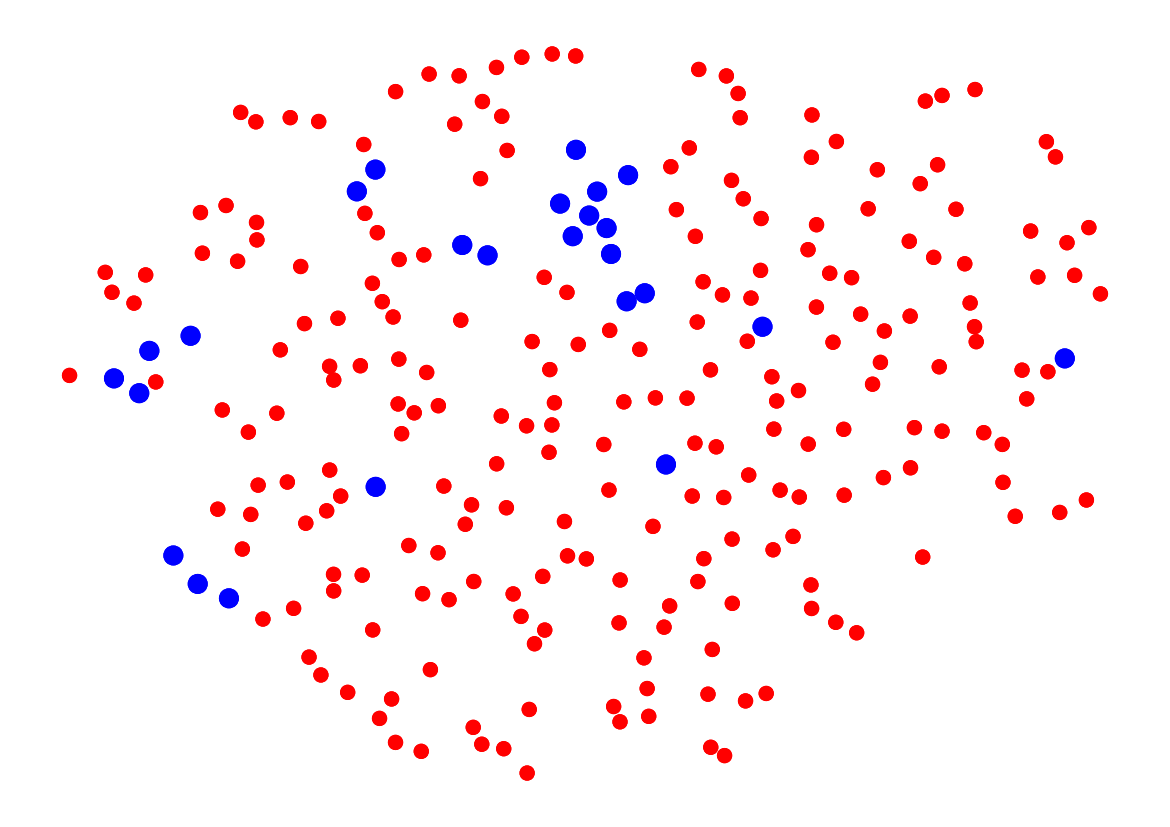} 
&\includegraphics[width=0.18\linewidth]{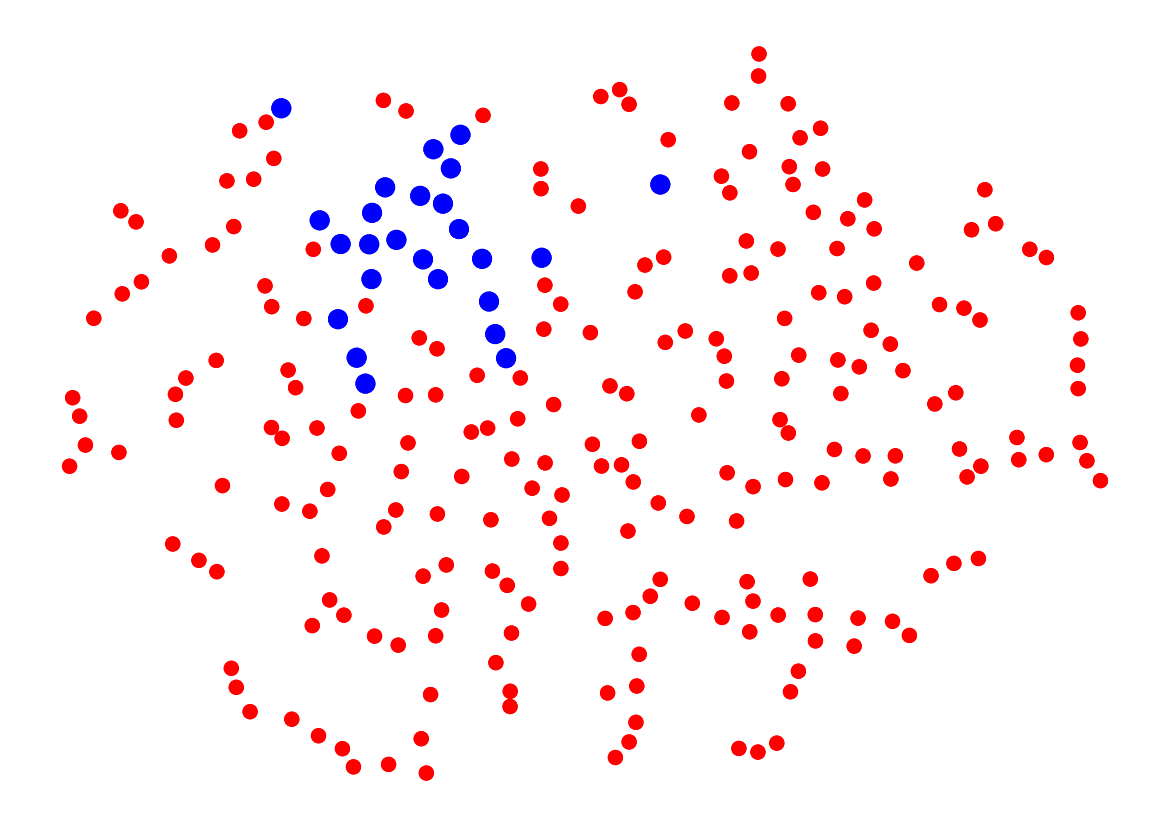} \\
(a) Original Graph  & (b)  NetGAN & (c) GAE & (d)  TagGen & (e) \name \\
\end{tabular}
\end{center}
\caption{TSNE Visualization of different synthetic graphs and the original graph.}
\label{fig:visualization}
\end{figure*}

\subsection{Visualization}
In this subsection, we visualize the synthetic graphs generated by deep learning baseline methods, including NetGAN, GAE, TagGen, and \name, as shown in Figure~\ref{fig:visualization}. To better visualize the graph, the experiment is conducted on the synthetic dataset shown in Figure 1. Here is the procedure to price the TSNE~\cite{van2008visualizing} visualization: we first use Node2vec to learn the representation for a graph (e.g., either the original graph or the synthetic graph), and then we use TSNE to reduce the representation to 2D space and visualize the graph. By observation, we can find that our proposed method can fairly preserve the topological structure for both the protected group and unprotected group while other baseline methods fail to do that.

\section{Related Work}

\textbf{Graph Generative Model. }
Graph generative models have a longstanding history, with applications in biology~\cite{ye2012exploring}, chemistry~\cite{you2018graphrnn,you2018graph}, and social sciences~\cite{ye2012exploring, DBLP:journals/fdata/ZhouZXH19}. Classic graph generators are often designed as network-property oriented models, which capture and reproduce one or more important structure properties, \eg, power-law degree distribution~\cite{erdos1959random,albert2002statistical}, small diameters~\cite{watts1998collective}, motif distribution~\cite{leskovec2010kronecker,purohit2018temporal}, and densification in graph evolution~\cite{fischer2014dynamic}. More recently, deep generative models
\cite{bojchevski2018netgan, you2018graphrnn, simonovsky2018graphvae,DBLP:conf/ijcai/YuCW0W21, DBLP:conf/kdd/LiuLLWS021, DBLP:conf/icml/LuoYJ21,DBLP:conf/icml/ChenHHRL21, DBLP:conf/kdd/GuoDZ21,DBLP:conf/www/GoyalJR20,DBLP:conf/kdd/HuangSLH19, DBLP:conf/www/ZenoFN21,DBLP:conf/kdd/GuoZQWSY20,DBLP:conf/nips/LiaoLSWHDUZ19}
for graphs have received much research interest. For example, 
in~\cite{simonovsky2018graphvae}, the authors propose a variational auto-encoders based framework named GraphVAE, which is designed to generate a number of small graphs and then employ a subgraph matching algorithm to assemble them into a complete graph with the same size as the original network;
in~\cite{DBLP:conf/icml/LuoYJ21}, the authors revisit the molecular graph generation problem and propose a discrete latent variable model to accommodate the discrete graph signals in reality. 
in~\cite{DBLP:conf/icml/ChenHHRL21}, the authors develop a probabilistic model to marginalize out the node orderings and estimate the joint likelihood of graphs. Though ~\cite{DBLP:conf/icml/LuoYJ21, DBLP:conf/icml/JinBJ18, you2018graph} utilize the label information to constrain the graph generation, these methods are only designed for molecular graph rather than general networks. Most of the existing works are predominately designed for producing general-purpose graphs and they overlook both the label information and fairness requirements. \\
\textbf{Fair Graph Mining. }
Despite the long-standing research on graphs, recent studies show multiple evidences~\cite{DBLP:conf/kdd/KangHMT20,DBLP:conf/icml/BoseH19,DBLP:conf/emnlp/FisherMPC20,DBLP:conf/www/WuCSHWW21} that many graph mining models are biased and may lead to harmful discrimination in downstream applications. To fill this gap, a surge of research has been conducted to amend the biased graph learning models to be fair or invariant regarding specific variables. Until now, the existing literature in fair graph mining can be roughly classified into two categories, \ie, \emph{group fairness on graphs}~\cite{DBLP:conf/icml/BoseH19, DBLP:conf/emnlp/FisherMPC20, DBLP:conf/www/Fu0MCBH23, DBLP:conf/www/WuCSHWW21, DBLP:conf/aaai/MasrourWYTE20, DBLP:conf/iclr/LiWZHL21, DBLP:conf/www/Tsioutsiouliklis21, DBLP:conf/icml/KleindessnerSAM19, DBLP:conf/icml/BuylB20, DBLP:conf/www/FarnadiBG20} and \emph{individual fairness on graphs}~\cite{DBLP:conf/kdd/DongKTL21, DBLP:journals/corr/abs-2105-03714, DBLP:conf/kdd/KangHMT20}. The former category aims to mitigate the bias and potential discrimination among demographic groups of nodes or edges in a wide spectrum of graph mining tasks, including graph proximity learning~\cite{DBLP:conf/www/Tsioutsiouliklis21}, graph clustering~\cite{DBLP:conf/icml/KleindessnerSAM19}, graph representation learning~\cite{DBLP:conf/icml/BoseH19,DBLP:conf/icml/BuylB20}. The latter category studies the problem of how to ensure the similar graph signals receive similar algorithmic outcomes~\cite{DBLP:conf/kdd/KangHMT20}. 
In this paper, we study the problem in the context of group fairness on graphs and make the initial effort to debiasing representation disparity in graph generative models. 

\section{Conclusion}
In this paper, we present \name\ - a novel generative model that incorporates the label information and fairness constraint in the graph generation process. \name\ is developed based on a self-paced learning paradigm that globally maintains a label-informed graph generation module and a fair learning module to extract graph context information. It is designed to gradually mitigate representation disparity by learning from the `easy' concepts to the `hard' ones to accurately capture the behavior of the protected groups and unprotected groups. The experimental results demonstrate the effectiveness of \name\ 
in generating high-quality graphs, alleviating the representation disparity, and enabling effective data augmentation for downstream applications.

\section*{Acknowledgment}

\thanks{This work is supported by NSF(1939725, 2137468), by the United States Air Force and DARPA under contract number FA8750-17-C-0153~\footnote{ Distribution Statement "A" (Approved for Public Release, Distribution Unlimited)}. The content of the information in this document does not necessarily reflect the position or the policy of the Government or Amazon, and no official endorsement should be inferred.  The U.S. Government is authorized to reproduce and distribute reprints for Government purposes notwithstanding any copyright notation here on.

}
\bibliographystyle{IEEEtran}
\bibliography{sample-base}

\begin{thebibliography}{10}
\providecommand{\url}[1]{#1}
\csname url@samestyle\endcsname
\providecommand{\newblock}{\relax}
\providecommand{\bibinfo}[2]{#2}
\providecommand{\BIBentrySTDinterwordspacing}{\spaceskip=0pt\relax}
\providecommand{\BIBentryALTinterwordstretchfactor}{4}
\providecommand{\BIBentryALTinterwordspacing}{\spaceskip=\fontdimen2\font plus
\BIBentryALTinterwordstretchfactor\fontdimen3\font minus
  \fontdimen4\font\relax}
\providecommand{\BIBforeignlanguage}[2]{{%
\expandafter\ifx\csname l@#1\endcsname\relax
\typeout{** WARNING: IEEEtran.bst: No hyphenation pattern has been}%
\typeout{** loaded for the language `#1'. Using the pattern for}%
\typeout{** the default language instead.}%
\else
\language=\csname l@#1\endcsname
\fi
#2}}
\providecommand{\BIBdecl}{\relax}
\BIBdecl

\bibitem{chakrabarti2006graph}
D.~Chakrabarti and C.~Faloutsos, ``Graph mining: Laws, generators, and
  algorithms,'' \emph{{ACM} Comput. Surv.}, vol.~38, no.~1, p.~2, 2006.

\bibitem{akoglu2008rtm}
L.~Akoglu, M.~McGlohon, and C.~Faloutsos, ``{RTM:} laws and a recursive
  generator for weighted time-evolving graphs,'' in \emph{Proceedings of the
  8th {IEEE} International Conference on Data Mining {(ICDM} 2008), December
  15-19, 2008, Pisa, Italy}.\hskip 1em plus 0.5em minus 0.4em\relax {IEEE}
  Computer Society, 2008, pp. 701--706.

\bibitem{stokes2020deep}
J.~M. Stokes, K.~Yang, K.~Swanson, W.~Jin, A.~Cubillos-Ruiz, N.~M. Donghia,
  C.~R. MacNair, S.~French, L.~A. Carfrae, Z.~Bloom-Ackerman \emph{et~al.}, ``A
  deep learning approach to antibiotic discovery,'' \emph{Cell}, 2020.

\bibitem{DBLP:conf/icml/JinBJ18}
W.~Jin, R.~Barzilay, and T.~S. Jaakkola, ``Junction tree variational
  autoencoder for molecular graph generation,'' in \emph{Proceedings of the
  35th International Conference on Machine Learning, {ICML} 2018,
  Stockholmsm{\"{a}}ssan, Stockholm, Sweden, July 10-15, 2018}, ser.
  Proceedings of Machine Learning Research, vol.~80.\hskip 1em plus 0.5em minus
  0.4em\relax {PMLR}, 2018, pp. 2328--2337.

\bibitem{bojchevski2018netgan}
A.~Bojchevski, O.~Shchur, D.~Z{\"{u}}gner, and S.~G{\"{u}}nnemann, ``Netgan:
  Generating graphs via random walks,'' in \emph{Proceedings of the 35th
  International Conference on Machine Learning, {ICML} 2018,
  Stockholmsm{\"{a}}ssan, Stockholm, Sweden, July 10-15, 2018}, ser.
  Proceedings of Machine Learning Research, vol.~80.\hskip 1em plus 0.5em minus
  0.4em\relax {PMLR}, 2018, pp. 609--618.

\bibitem{albert2002statistical}
R.~Albert and A.-L. Barab{\'a}si, ``Statistical mechanics of complex
  networks,'' \emph{Reviews of modern physics}, 2002.

\bibitem{akoglu2009rtg}
L.~Akoglu and C.~Faloutsos, ``{RTG:} {A} recursive realistic graph generator
  using random typing,'' in \emph{Machine Learning and Knowledge Discovery in
  Databases, European Conference, {ECML} {PKDD} 2009, Bled, Slovenia, September
  7-11, 2009, Proceedings, Part {I}}, ser. Lecture Notes in Computer Science,
  vol. 5781.\hskip 1em plus 0.5em minus 0.4em\relax Springer, 2009, pp. 13--28.

\bibitem{leskovec2010kronecker}
J.~Leskovec, D.~Chakrabarti, J.~M. Kleinberg, C.~Faloutsos, and Z.~Ghahramani,
  ``Kronecker graphs: An approach to modeling networks,'' \emph{J. Mach. Learn.
  Res.}, vol.~11, pp. 985--1042, 2010.

\bibitem{kim2012multiplicative}
M.~Kim and J.~Leskovec, ``Multiplicative attribute graph model of real-world
  networks,'' \emph{Internet Math.}, 2012.

\bibitem{watts1998collective}
C.~Grabow, S.~Grosskinsky, J.~Kurths, and M.~Timme, ``Collective relaxation
  dynamics of small-world networks,'' \emph{Physical Review E}, vol.~91, no.~5,
  p. 052815, 2015.

\bibitem{fischer2014dynamic}
F.~Fischer and C.~Helmberg, ``Dynamic graph generation for the shortest path
  problem in time expanded networks,'' \emph{Math. Program.}, vol. 143, no.
  1-2, pp. 257--297, 2014.

\bibitem{waxman1988routing}
B.~M. Waxman, ``Routing of multipoint connections,'' \emph{{IEEE} J. Sel. Areas
  Commun.}, vol.~6, no.~9, pp. 1617--1622, 1988.

\bibitem{jing2023sterling}
B.~Jing, Y.~Yan, K.~Ding, C.~Park, Y.~Zhu, H.~Liu, and H.~Tong, ``Sterling:
  Synergistic representation learning on bipartite graphs,'' \emph{arXiv
  preprint arXiv:2302.05428}, 2023.

\bibitem{jing2022coin}
B.~Jing, Y.~Yan, Y.~Zhu, and H.~Tong, ``Coin: Co-cluster infomax for bipartite
  graphs,'' \emph{arXiv preprint arXiv:2206.00006}, 2022.

\bibitem{zhao2011synchronization}
L.~Zhao, B.~B. II, T.~I. Netoff, and D.~Q. Nykamp, ``Synchronization from
  second order network connectivity statistics,'' \emph{Frontiers Comput.
  Neurosci.}, vol.~5, p.~28, 2011.

\bibitem{zheng2021deeper}
L.~Zheng, D.~Fu, R.~Maciejewski, and J.~He, ``Deeper-gxx: deepening arbitrary
  gnns,'' \emph{arXiv preprint arXiv:2110.13798}, 2021.

\bibitem{DBLP:conf/cikm/ZhouZF0H22}
D.~Zhou, L.~Zheng, D.~Fu, J.~Han, and J.~He, ``Mentorgnn: Deriving curriculum
  for pre-training gnns,'' in \emph{Proceedings of the 31st {ACM} International
  Conference on Information {\&} Knowledge Management, Atlanta, GA, USA,
  October 17-21, 2022}, M.~A. Hasan and L.~Xiong, Eds.\hskip 1em plus 0.5em
  minus 0.4em\relax {ACM}, 2022, pp. 2721--2731.

\bibitem{jing2021hdmi}
B.~Jing, C.~Park, and H.~Tong, ``Hdmi: High-order deep multiplex infomax,'' in
  \emph{Proceedings of the Web Conference 2021}, 2021, pp. 2414--2424.

\bibitem{jing2021multiplex}
B.~Jing, Z.~You, T.~Yang, W.~Fan, and H.~Tong, ``Multiplex graph neural network
  for extractive text summarization,'' in \emph{Proceedings of the 2021
  Conference on Empirical Methods in Natural Language Processing}, 2021, pp.
  133--139.

\bibitem{you2018graphrnn}
J.~You, R.~Ying, X.~Ren, W.~L. Hamilton, and J.~Leskovec, ``Graphrnn:
  Generating realistic graphs with deep auto-regressive models,'' in
  \emph{Proceedings of the 35th International Conference on Machine Learning,
  {ICML} 2018, Stockholmsm{\"{a}}ssan, Stockholm, Sweden, July 10-15, 2018},
  ser. Proceedings of Machine Learning Research, vol.~80.\hskip 1em plus 0.5em
  minus 0.4em\relax {PMLR}, 2018, pp. 5694--5703.

\bibitem{simonovsky2018graphvae}
M.~Simonovsky and N.~Komodakis, ``Graphvae: Towards generation of small graphs
  using variational autoencoders,'' pp. 412--422, 2018.

\bibitem{guo2020systematic}
X.~Guo and L.~Zhao, ``A systematic survey on deep generative models for graph
  generation,'' \emph{{IEEE} Trans. Pattern Anal. Mach. Intell.}, vol.~45,
  no.~5, pp. 5370--5390, 2023.

\bibitem{li2018learning}
Y.~Li, O.~Vinyals, C.~Dyer, R.~Pascanu, and P.~W. Battaglia, ``Learning deep
  generative models of graphs,'' \emph{CoRR}, vol. abs/1803.03324, 2018.

\bibitem{grover2019graphite}
A.~Grover, A.~Zweig, and S.~Ermon, ``Graphite: Iterative generative modeling of
  graphs,'' in \emph{ICML}.\hskip 1em plus 0.5em minus 0.4em\relax PMLR, 2019,
  pp. 2434--2444.

\bibitem{DBLP:conf/www/GoyalJR20}
N.~Goyal, H.~V. Jain, and S.~Ranu, ``Graphgen: {A} scalable approach to
  domain-agnostic labeled graph generation,'' in \emph{{WWW} '20: The Web
  Conference 2020}.\hskip 1em plus 0.5em minus 0.4em\relax {ACM} / {IW3C2},
  2020, pp. 1253--1263.

\bibitem{harrison2009identity}
R.~Harrison and M.~Thomas, ``Identity in online communities: Social networking
  sites and language learning,'' \emph{International Journal of Emerging
  Technologies and Society}, 2009.

\bibitem{wellman1999network}
B.~Wellman, ``The network community: An introduction,'' \emph{Networks in the
  global village}, 1999.

\bibitem{gajane2017formalizing}
\BIBentryALTinterwordspacing
P.~Gajane, ``On formalizing fairness in prediction with machine learning,''
  \emph{CoRR}, vol. abs/1710.03184, 2017. [Online]. Available:
  \url{http://arxiv.org/abs/1710.03184}
\BIBentrySTDinterwordspacing

\bibitem{DBLP:conf/sdm/ZhengZH23}
L.~Zheng, Y.~Zhu, and J.~He, ``Fairness-aware multi-view clustering,'' in
  \emph{Proceedings of the 2023 {SIAM} International Conference on Data Mining,
  {SDM} 2023, Minneapolis-St. Paul Twin Cities, MN, USA, April 27-29, 2023},
  S.~Shekhar, Z.~Zhou, Y.~Chiang, and G.~Stiglic, Eds.\hskip 1em plus 0.5em
  minus 0.4em\relax {SIAM}, 2023, pp. 856--864.

\bibitem{DBLP:journals/corr/abs-1908-09635}
N.~Mehrabi, F.~Morstatter, N.~Saxena, K.~Lerman, and A.~Galstyan, ``A survey on
  bias and fairness in machine learning,'' \emph{CoRR}, 2019.

\bibitem{hashimoto2018fairness}
T.~B. Hashimoto, M.~Srivastava, H.~Namkoong, and P.~Liang, ``Fairness without
  demographics in repeated loss minimization,'' in \emph{Proceedings of the
  35th International Conference on Machine Learning, {ICML} 2018,
  Stockholmsm{\"{a}}ssan, Stockholm, Sweden, July 10-15, 2018}, ser.
  Proceedings of Machine Learning Research, vol.~80.\hskip 1em plus 0.5em minus
  0.4em\relax {PMLR}, 2018, pp. 1934--1943.

\bibitem{mikolov2010recurrent}
T.~Mikolov, M.~Karafi{\'{a}}t, L.~Burget, J.~Cernock{\'{y}}, and S.~Khudanpur,
  ``Recurrent neural network based language model,'' in \emph{{INTERSPEECH}
  2010, 11th Annual Conference of the International Speech Communication
  Association, Makuhari, Chiba, Japan, September 26-30, 2010}.\hskip 1em plus
  0.5em minus 0.4em\relax {ISCA}, 2010, pp. 1045--1048.

\bibitem{hochreiter1997long}
S.~Hochreiter and J.~Schmidhuber, ``Long short-term memory,'' \emph{Neural
  Comput.}, vol.~9, no.~8, pp. 1735--1780, 1997.

\bibitem{DBLP:conf/icml/BoseH19}
A.~J. Bose and W.~L. Hamilton, ``Compositional fairness constraints for graph
  embeddings,'' in \emph{Proceedings of the 36th ICML}, vol.~97.\hskip 1em plus
  0.5em minus 0.4em\relax {PMLR}, 2019, pp. 715--724.

\bibitem{zemel2013learning}
R.~S. Zemel, Y.~Wu, K.~Swersky, T.~Pitassi, and C.~Dwork, ``Learning fair
  representations,'' in \emph{Proceedings of the 30th International Conference
  on Machine Learning, {ICML} 2013, Atlanta, GA, USA, 16-21 June 2013}, ser.
  {JMLR} Workshop and Conference Proceedings, vol.~28.\hskip 1em plus 0.5em
  minus 0.4em\relax JMLR.org, 2013, pp. 325--333.

\bibitem{vaswani2017attention}
A.~Vaswani, N.~Shazeer, N.~Parmar, J.~Uszkoreit, L.~Jones, A.~N. Gomez,
  L.~Kaiser, and I.~Polosukhin, ``Attention is all you need,'' in
  \emph{Advances in Neural Information Processing Systems 2017, December 4-9,
  2017, Long Beach, CA, {USA}}, 2017, pp. 5998--6008.

\bibitem{apers2019expansion}
S.~Apers, ``Expansion testing using quantum fast-forwarding and seed sets,''
  \emph{Quantum}, vol.~4, p. 323, 2020.

\bibitem{spielman2013local}
D.~A. Spielman and S.~Teng, ``A local clustering algorithm for massive graphs
  and its application to nearly linear time graph partitioning,'' \emph{{SIAM}
  J. Comput.}, 2013.

\bibitem{grover2016node2vec}
A.~Grover and J.~Leskovec, ``node2vec: Scalable feature learning for
  networks,'' in \emph{Proceedings of the 22nd {ACM} {SIGKDD} International
  Conference on Knowledge Discovery and Data Mining, San Francisco, CA, USA,
  August 13-17, 2016}.\hskip 1em plus 0.5em minus 0.4em\relax {ACM}, 2016, pp.
  855--864.

\bibitem{mikolov2013distributed}
T.~Mikolov, I.~Sutskever, K.~Chen, G.~S. Corrado, and J.~Dean, ``Distributed
  representations of words and phrases and their compositionality,'' in
  \emph{Advances in Neural Information Processing Systems 2013. Proceedings of
  a meeting held December 5-8, 2013, Lake Tahoe, Nevada, United States}, 2013,
  pp. 3111--3119.

\bibitem{mikolov2013efficient}
T.~Mikolov, K.~Chen, G.~Corrado, and J.~Dean, ``Efficient estimation of word
  representations in vector space,'' in \emph{1st International Conference on
  Learning Representations, {ICLR} 2013, Scottsdale, Arizona, USA, May 2-4,
  2013, Workshop Track Proceedings}, 2013.

\bibitem{kumar2010self}
M.~P. Kumar, B.~Packer, and D.~Koller, ``Self-paced learning for latent
  variable models,'' in \emph{Advances in 24th Annual Conference on Neural
  Information Processing Systems 2010. Proceedings of a meeting held 6-9
  December 2010, Vancouver, British Columbia, Canada}.\hskip 1em plus 0.5em
  minus 0.4em\relax Curran Associates, Inc., 2010, pp. 1189--1197.

\bibitem{bottou2010large}
L.~Bottou, ``Large-scale machine learning with stochastic gradient descent,''
  in \emph{19th International Conference on Computational Statistics,
  {COMPSTAT} 2010, Paris, France, August 22-27, 2010 - Keynote, Invited and
  Contributed Papers}.\hskip 1em plus 0.5em minus 0.4em\relax Physica-Verlag,
  2010, pp. 177--186.

\bibitem{leskovec2015snap}
J.~Leskovec and R.~Sosi{\v{c}}, ``Snap: A general-purpose network analysis and
  graph-mining library,'' \emph{ACM Transactions on Intelligent Systems and
  Technology (TIST)}, vol.~8, no.~1, pp. 1--20, 2016.

\bibitem{tang2009relational}
L.~Tang and H.~Liu, ``Relational learning via latent social dimensions,'' in
  \emph{Proceedings of the 15th {ACM} {SIGKDD} International Conference on
  Knowledge Discovery and Data Mining, Paris, France, June 28 - July 1,
  2009}.\hskip 1em plus 0.5em minus 0.4em\relax {ACM}, 2009, pp. 817--826.

\bibitem{ding2019interactive}
K.~Ding, J.~Li, and H.~Liu, ``Interactive anomaly detection on attributed
  networks,'' in \emph{Proceedings of the Twelfth {ACM} International
  Conference on Web Search and Data Mining, {WSDM} 2019, Melbourne, VIC,
  Australia, February 11-15, 2019}.\hskip 1em plus 0.5em minus 0.4em\relax
  {ACM}, 2019, pp. 357--365.

\bibitem{erdos1959random}
P.~ERDdS and A.~R\&wi, ``On random graphs i,'' \emph{Publ. math. debrecen},
  1959.

\bibitem{kipf2016variational}
T.~N. Kipf and M.~Welling, ``Variational graph auto-encoders,'' \emph{arXiv
  preprint arXiv:1611.07308}, 2016.

\bibitem{DBLP:conf/kdd/ZhouZ0H20}
D.~Zhou, L.~Zheng, J.~Han, and J.~He, ``A data-driven graph generative model
  for temporal interaction networks,'' in \emph{{KDD} '20: The 26th {ACM}
  {SIGKDD} Conference on Knowledge Discovery and Data Mining, Virtual Event,
  CA, USA, August 23-27, 2020}.\hskip 1em plus 0.5em minus 0.4em\relax {ACM},
  2020, pp. 401--411.

\bibitem{pearce2005improved}
D.~J. Pearce, ``An improved algorithm for finding the strongly connected
  components of a directed graph,'' \emph{Victoria University, Wellington, NZ,
  Tech. Rep}, 2005.

\bibitem{van2008visualizing}
L.~Van~der Maaten and G.~Hinton, ``Visualizing data using t-sne.''
  \emph{Journal of machine learning research}, vol.~9, no.~11, 2008.

\bibitem{ye2012exploring}
M.~Ye, X.~Liu, and W.~Lee, ``Exploring social influence for recommendation: a
  generative model approach,'' in \emph{The 35th International {ACM} {SIGIR}
  conference on research and development in Information Retrieval, {SIGIR} '12,
  Portland, OR, USA, August 12-16, 2012}.\hskip 1em plus 0.5em minus
  0.4em\relax {ACM}, 2012, pp. 671--680.

\bibitem{you2018graph}
J.~You, B.~Liu, Z.~Ying, V.~S. Pande, and J.~Leskovec, ``Graph convolutional
  policy network for goal-directed molecular graph generation,'' in
  \emph{Advances in Neural Information Processing Systems 2018, NeurIPS 2018,
  December 3-8, 2018, Montr{\'{e}}al, Canada}, 2018, pp. 6412--6422.

\bibitem{DBLP:journals/fdata/ZhouZXH19}
D.~Zhou, L.~Zheng, J.~Xu, and J.~He, ``Misc-gan: {A} multi-scale generative
  model for graphs,'' \emph{Frontiers Big Data}, vol.~2, p.~3, 2019.

\bibitem{purohit2018temporal}
S.~Purohit, L.~B. Holder, and G.~Chin, ``Temporal graph generation based on a
  distribution of temporal motifs,'' in \emph{Proceedings of the 14th
  International Workshop on Mining and Learning with Graphs}, vol.~7, 2018.

\bibitem{DBLP:conf/ijcai/YuCW0W21}
J.~Yu, Y.~Chai, Y.~Wang, Y.~Hu, and Q.~Wu, ``Cogtree: Cognition tree loss for
  unbiased scene graph generation,'' in \emph{Proceedings of the Thirtieth
  IJCAI}.\hskip 1em plus 0.5em minus 0.4em\relax ijcai.org, 2021, pp.
  1274--1280.

\bibitem{DBLP:conf/kdd/LiuLLWS021}
D.~Liu, J.~Lian, Z.~Liu, X.~Wang, G.~Sun, and X.~Xie, ``Reinforced anchor
  knowledge graph generation for news recommendation reasoning,'' in
  \emph{{KDD} '21: The 27th {ACM} {SIGKDD} 2021}.\hskip 1em plus 0.5em minus
  0.4em\relax {ACM}, 2021, pp. 1055--1065.

\bibitem{DBLP:conf/icml/LuoYJ21}
Y.~Luo, K.~Yan, and S.~Ji, ``Graphdf: {A} discrete flow model for molecular
  graph generation,'' in \emph{Proceedings of {ICML} 2021}, vol. 139.\hskip 1em
  plus 0.5em minus 0.4em\relax {PMLR}, 2021, pp. 7192--7203.

\bibitem{DBLP:conf/icml/ChenHHRL21}
X.~Chen, X.~Han, J.~Hu, F.~J.~R. Ruiz, and L.~Liu, ``Order matters:
  Probabilistic modeling of node sequence for graph generation,'' in
  \emph{Proceedings of the 38th ICML}, vol. 139.\hskip 1em plus 0.5em minus
  0.4em\relax {PMLR}, 2021, pp. 1630--1639.

\bibitem{DBLP:conf/kdd/GuoDZ21}
X.~Guo, Y.~Du, and L.~Zhao, ``Deep generative models for spatial networks,'' in
  \emph{{KDD} '21: The 27th {ACM} {SIGKDD} 2021}.\hskip 1em plus 0.5em minus
  0.4em\relax {ACM}, 2021, pp. 505--515.

\bibitem{DBLP:conf/kdd/HuangSLH19}
X.~Huang, Q.~Song, Y.~Li, and X.~Hu, ``Graph recurrent networks with attributed
  random walks,'' in \emph{Proceedings of the 25th {ACM} {SIGKDD}}.\hskip 1em
  plus 0.5em minus 0.4em\relax {ACM}, 2019, pp. 732--740.

\bibitem{DBLP:conf/www/ZenoFN21}
G.~Zeno, T.~L. Fond, and J.~Neville, ``{DYMOND:} dynamic motif-nodes network
  generative model,'' in \emph{{WWW} '21: The Web Conference 2021}.\hskip 1em
  plus 0.5em minus 0.4em\relax {ACM} / {IW3C2}, 2021, pp. 718--729.

\bibitem{DBLP:conf/kdd/GuoZQWSY20}
X.~Guo, L.~Zhao, Z.~Qin, L.~Wu, A.~Shehu, and Y.~Ye, ``Interpretable deep graph
  generation with node-edge co-disentanglement,'' in \emph{{KDD} '20: The 26th
  {ACM} {SIGKDD} 2020}.\hskip 1em plus 0.5em minus 0.4em\relax {ACM}, 2020, pp.
  1697--1707.

\bibitem{DBLP:conf/nips/LiaoLSWHDUZ19}
R.~Liao, Y.~Li, Y.~Song, S.~Wang, W.~L. Hamilton, D.~Duvenaud, R.~Urtasun, and
  R.~S. Zemel, ``Efficient graph generation with graph recurrent attention
  networks,'' in \emph{Advances in Neural Information Processing Systems 2019,
  NeurIPS 2019, December 8-14, 2019, Vancouver, BC, Canada}, 2019, pp.
  4257--4267.

\bibitem{DBLP:conf/kdd/KangHMT20}
J.~Kang, J.~He, R.~Maciejewski, and H.~Tong, ``Inform: Individual fairness on
  graph mining,'' in \emph{{KDD} '20: The 26th {ACM} {SIGKDD} 2020}.\hskip 1em
  plus 0.5em minus 0.4em\relax {ACM}, 2020, pp. 379--389.

\bibitem{DBLP:conf/emnlp/FisherMPC20}
J.~Fisher, A.~Mittal, D.~Palfrey, and C.~Christodoulopoulos, ``Debiasing
  knowledge graph embeddings,'' in \emph{Proceedings of the 2020 Conference on
  EMNLP 2020}.\hskip 1em plus 0.5em minus 0.4em\relax Association for
  Computational Linguistics, 2020, pp. 7332--7345.

\bibitem{DBLP:conf/www/WuCSHWW21}
L.~Wu, L.~Chen, P.~Shao, R.~Hong, X.~Wang, and M.~Wang, ``Learning fair
  representations for recommendation: {A} graph-based perspective,'' in
  \emph{{WWW} '21: The Web Conference 2021}.\hskip 1em plus 0.5em minus
  0.4em\relax {ACM} / {IW3C2}, 2021, pp. 2198--2208.

\bibitem{DBLP:conf/www/Fu0MCBH23}
D.~Fu, D.~Zhou, R.~Maciejewski, A.~Croitoru, M.~Boyd, and J.~He,
  ``Fairness-aware clique-preserving spectral clustering of temporal graphs,''
  in \emph{Proceedings of the {ACM} Web Conference 2023, {WWW} 2023, Austin,
  TX, USA, 30 April 2023 - 4 May 2023}.\hskip 1em plus 0.5em minus 0.4em\relax
  {ACM}, 2023, pp. 3755--3765.

\bibitem{DBLP:conf/aaai/MasrourWYTE20}
F.~Masrour, T.~Wilson, H.~Yan, P.~Tan, and A.~Esfahanian, ``Bursting the filter
  bubble: Fairness-aware network link prediction,'' in \emph{The Thirty-Fourth
  AAAI}.\hskip 1em plus 0.5em minus 0.4em\relax {AAAI} Press, 2020, pp.
  841--848.

\bibitem{DBLP:conf/iclr/LiWZHL21}
P.~Li, Y.~Wang, H.~Zhao, P.~Hong, and H.~Liu, ``On dyadic fairness: Exploring
  and mitigating bias in graph connections,'' in \emph{{ICLR} 2021}.\hskip 1em
  plus 0.5em minus 0.4em\relax OpenReview.net, 2021.

\bibitem{DBLP:conf/www/Tsioutsiouliklis21}
S.~Tsioutsiouliklis, E.~Pitoura, P.~Tsaparas, I.~Kleftakis, and N.~Mamoulis,
  ``Fairness-aware pagerank,'' in \emph{{WWW} '21: The Web Conference
  2021}.\hskip 1em plus 0.5em minus 0.4em\relax {ACM} / {IW3C2}, 2021, pp.
  3815--3826.

\bibitem{DBLP:conf/icml/KleindessnerSAM19}
M.~Kleindessner, S.~Samadi, P.~Awasthi, and J.~Morgenstern, ``Guarantees for
  spectral clustering with fairness constraints,'' in \emph{{ICML} 2019, 9-15
  June 2019, Long Beach, California, {USA}}, vol.~97.\hskip 1em plus 0.5em
  minus 0.4em\relax {PMLR}, 2019, pp. 3458--3467.

\bibitem{DBLP:conf/icml/BuylB20}
M.~Buyl and T.~D. Bie, ``Debayes: a bayesian method for debiasing network
  embeddings,'' in \emph{Proceedings of the 37th ICML}, vol. 119.\hskip 1em
  plus 0.5em minus 0.4em\relax {PMLR}, 2020, pp. 1220--1229.

\bibitem{DBLP:conf/www/FarnadiBG20}
G.~Farnadi, B.~Babaki, and M.~Gendreau, ``A unifying framework for
  fairness-aware influence maximization,'' in \emph{Companion of The 2020 Web
  Conference 2020}.\hskip 1em plus 0.5em minus 0.4em\relax {ACM}, 2020, pp.
  714--722.

\bibitem{DBLP:conf/kdd/DongKTL21}
Y.~Dong, J.~Kang, H.~Tong, and J.~Li, ``Individual fairness for graph neural
  networks: {A} ranking based approach,'' in \emph{{KDD} '21: The 27th {ACM}
  {SIGKDD} 2021}.\hskip 1em plus 0.5em minus 0.4em\relax {ACM}, 2021, pp.
  300--310.

\bibitem{DBLP:journals/corr/abs-2105-03714}
S.~Gupta and A.~Dukkipati, ``Protecting individual interests across clusters:
  Spectral clustering with guarantees,'' \emph{CoRR}, vol. abs/2105.03714,
  2021.

\end{thebibliography}

\end{document}